\documentclass[11pt]{article}

\usepackage[preprint]{acl}
\usepackage{multirow}
\usepackage{graphicx}
\usepackage[table,xcdraw]{xcolor}
\usepackage[normalem]{ulem}
\usepackage{float}
\useunder{\uline}{\ul}{}
\usepackage{times}
\usepackage{latexsym}
\usepackage{amssymb}
\usepackage{url}
\usepackage{hyperref}
\usepackage{url}
\usepackage{hyperref}

\usepackage[T1]{fontenc}
\usepackage[utf8]{inputenc}

\usepackage{microtype}

\usepackage{inconsolata}

\usepackage{booktabs}
\usepackage{comment}

\usepackage{adjustbox}
\usepackage{rotating}

\usepackage[most]{tcolorbox}
\usepackage{listings}
\usepackage{wrapfig,lipsum,booktabs}

\definecolor{codebg}{RGB}{245,245,245}
\definecolor{codeframe}{RGB}{220,220,220}
\definecolor{codetitle}{RGB}{80,80,80}
\definecolor{keycolor}{RGB}{0,0,255}
\definecolor{stringcolor}{RGB}{0,128,0}
\definecolor{commentcolor}{RGB}{128,128,128}

\lstdefinelanguage{yaml}{
  keywords={true,false,null,y,n},
  keywordstyle=\color{keycolor}\bfseries,
  basicstyle=\ttfamily\footnotesize,
  breaklines=true,
  comment=[l]{\#},
  commentstyle=\color{commentcolor}\itshape,
  stringstyle=\color{stringcolor},
  moredelim=[l][\color{orange}]{\&},
  moredelim=[l][\color{orange}]{\*},
  morestring=[b]',
  morestring=[b]",
  literate={-}{{{\color{red}-}}}1 {:}{{{\color{red}:}}}1,
}

\newtcolorbox{codeblock}[2][]{
    enhanced,
    boxrule=0.5pt,
    colback=codebg,
    colframe=codeframe,
    coltitle=codetitle,
    fonttitle=\bfseries\ttfamily\footnotesize,
    title={#2},
    attach boxed title to top left={yshift=-2mm, xshift=2mm},
    boxed title style={colback=white, boxrule=0.5pt, colframe=codeframe},
    arc=2pt,
    breakable,
    #1
}

\usepackage{siunitx}
\usepackage{xfp}
\usepackage{graphicx}

\usepackage{xcolor}
\usepackage{enumitem}
\title{AfriqueLLM: How Data Mixing and Model Architecture Impact Continued Pre-training for African Languages}

\author{
  \textbf{Hao Yu\textsuperscript{1,2}},
  \textbf{Tianyi Xu\textsuperscript{1,2}},
  \textbf{Michael A. Hedderich\textsuperscript{3}},
\\
  \textbf{Wassim Hamidouche\textsuperscript{4}},
  \textbf{Syed Waqas Zamir\textsuperscript{4}},
  \textbf{David Ifeoluwa Adelani\textsuperscript{1,2,5}}
\\
\\
  \textsuperscript{1}McGill University, Canada,
  \textsuperscript{2}Mila-Quebec AI Institute, Canada,
\\
  \textsuperscript{3}LMU Munich \& Munich Center for Machine Learning, Germany,
\\
  \textsuperscript{4}Microsoft AI for Good Research Lab,
  \textsuperscript{5}Canada CIFAR AI Chair
\\
  \small{
    \textbf{Correspondence:} \href{mailto:hao.yu2@mail.mcgill.ca}{hao.yu2@mail.mcgill.ca}, \href{mailto:david.adelani@mila.quebec}{david.adelani@mila.quebec}
  }
}

\begin{document}
\maketitle
\begin{abstract}
Large language models (LLMs) are increasingly multilingual, yet open models continue to underperform relative to proprietary systems, with the gap most pronounced for African languages. Continued pre-training (CPT) offers a practical route to language adaptation, but improvements on demanding capabilities such as mathematical reasoning often remain limited. This limitation is driven in part by the uneven domain coverage and missing task-relevant knowledge that characterize many low-resource language corpora. We present \texttt{AfriqueLLM}, a suite of open LLMs adapted to 20 African languages through CPT on 26B tokens. We perform a comprehensive empirical study across five base models spanning sizes and architectures, including Llama 3.1, Gemma 3, and Qwen 3, and systematically analyze how CPT data composition shapes downstream performance. In particular, we vary mixtures that include math, code, and synthetic translated data, and evaluate the resulting models on a range of multilingual benchmarks. Our results identify data composition as the primary driver of CPT gains. Adding math, code, and synthetic translated data yields consistent improvements, including on reasoning-oriented evaluations. Within a fixed architecture, larger models typically improve performance, but architectural choices dominate scale when comparing across model families. Moreover, strong multilingual performance in the base model does not reliably predict post-CPT outcomes; robust architectures coupled with task-aligned data provide a more dependable recipe. Finally, our best models improve long-context performance, including document-level translation. Models and code have been released on Huggingface and Github. \footnote{Model: \href{https://huggingface.co/collections/McGill-NLP/afriquellm}{AfriqueLLM Huggingface Collection}}\footnote{Code: \href{https://github.com/McGill-NLP/AfriqueLLM}{McGill-NLP/AfriqueLLM}} %Paper has been accepted to ACL 2026.
%\footnote{Accepted to ACL 2026.}

\end{abstract}

\section{Introduction}

Large language models (LLMs) are becoming increasingly multilingual, with proprietary models pre-trained on hundreds of languages~\citep{jaech2024openai,comanici2025gemini}. Open models follow a similar trend, but the performance gap with proprietary LLMs is often larger for low-resource languages, particularly African languages~\citep{adelani-etal-2025-irokobench,adebara-etal-2025-evaluating}. This gap highlights an opportunity to develop language- or region-specific LLMs for these languages.

Since the advent of pretrained language models such as BERT~\cite{devlin-etal-2019-bert}, continued pre-training has become a standard approach for adapting models to new domains and languages~\citep{gururangan-etal-2020-dont,chau-smith-2021-specializing,AfroXLM}, and has recently been scaled to modern LLMs~\citep{nguyen-etal-2024-seallms,ji2025massively,buzaaba2025lugha}. While CPT often yields significant improvements for natural language understanding (NLU) and translation tasks, gains on more challenging tasks, such as mathematical reasoning or knowledge-based QA (e.g., MMLU~\citep{hendrycks2021measuring}), remain limited due to uneven knowledge coverage across languages, with low-resource languages often spanning fewer domains~\citep{buzaaba2025lugha}.

To further improve downstream performance, LLMs are increasingly trained on heterogeneous data sources such as math, code, and other knowledge-rich corpora. These sources are often scarce in low-resource languages, yet they can substantially boost performance across a wide range of downstream tasks~\citep{aryabumi2024code,smollm3,li2025rethinking}. Recent work also shows that multilingual capability can be improved by training on machine-translated English data covering diverse domains, achieving competitive results even without adding monolingual data in the target languages~\citep{wang-etal-2025-multilingual-language}. Despite these advances, we still lack a comprehensive empirical understanding of how incorporating heterogeneous sources affects CPT outcomes for low-resource languages. In this work, we address this gap by systematically studying CPT data mixtures and analyzing how base-model architecture and prior language coverage influence downstream performance after adaptation.

%In this paper, we introduce \textbf{AfriqueLLM}, a suite of open language models adapted for 20 African languages through efficient CPT on 26B tokens, achieving state-of-the-art performance on $<15B$ parameter size models across multilingual benchmarks while maintaining English proficiency. We performed CPT on five base models from different architectures and model sizes such as Llama 3.1 8B, Gemma 3 (4B \& 12B), and Qwen 3 (8B \& 14B), and vary data mixture to improve performance on downstream tasks. 

We introduce \textbf{AfriqueLLM}, a suite of open language models adapted to 20 African languages via efficient continued pre-training (CPT) on 26B tokens. We perform CPT on several base model spanning different architectures and scales, including Llama 3.1 8B, Gemma 3 (4B and 12B), and Qwen 3 (4B, 8B, and 14B). Across these backbones, we systematically vary the CPT data mixture to quantify its impact on downstream performance. AfriqueLLM achieves strong results on multilingual benchmarks for models with fewer than 15B parameters, while largely preserving English performance.

Our evaluation leads to four main findings. \textit{(1)} The CPT data mixture is the strongest determinant of gains. Adding math, code, and synthetic translated data consistently improves performance. \textit{(2)} Within a fixed architecture, larger models generally perform better. Across architectures, however, scale alone is not predictive; for example, CPT-adapted Qwen 3 8B is competitive with Gemma 3 12B. \textit{(3)} Strong multilingual proficiency of the base model does not reliably translate into better post-CPT results. Instead, architectural choices and task-aligned data are more predictive. \textit{(4)} Our best models, Qwen 3 (8B and 14B), better preserve performance in high-resource languages after CPT and achieve strong results on long-context tasks such as document-level translation.

We hope these findings inform more effective adaptation of LLMs to low-resource languages. To support future work, we have publicly released all our CPT-adapted AfriqueLLMs on HuggingFace Model Hub.~\footnote{\href{https://huggingface.co/collections/McGill-NLP/afriquellm}{AfriqueLLM Collection}}

% \section{Background}
\section{Related Work}
\label{sec:related_work}

The landscape of LLMs has undergone a paradigm shift from model-centric architectures to data-centric methodologies. While early foundational work focused on scaling parameters and compute \cite{Kaplan2020,Brown2020}, recent advancements in 2024 and 2025 have demonstrated that data quality, mixture ratios, and curriculum learning are the primary drivers of performance \cite{llama3,Qwen3,smollm3,Olmo3,Nemotron3}. This section focuses on two important aspects of pre-training: (1) Data mixture and (2) Continued pre-training for low-resource languages.
\subsection{Data Quality, Mixture, and Synthetic Data}
\paragraph{Data Quality and Curation.}
Recent efforts have focused on improving the quality of web collected data such as FineWeb \cite{FineWeb} dataset in the English setting.  
In multilingual settings, FineWeb2 \cite{FineWeb2} extends these pipelines to scale pre-training data processing to over 1,000 languages. 
In the African context, this focus on quality has led to the creation of specialized datasets, such as WURA \cite{WURA} and MADLAD-400 \cite{MADLAD-400}.

\paragraph{Data Mixture and Ratios}
The importance of dynamic data mixtures is exemplified by the training recipes of recent models like SmolLM2~\cite{SmolLM2} and SmolLM3~\cite{smollm3}, which utilize multi-stage training curricula that adjust the ratio of web, code, and math data over time. OLMo2 and OLMo3 \citet{OLMo2,Olmo3} further validate this approach by introducing specialized data mixes (e.g., Dolmino Mix) during the annealing phase.
The scarcity of high-quality natural text for reasoning and low-resource languages has driven the adoption of synthetic data. \citet{Joshi2024} and \citet{Nemotron-4-340B} demonstrate that synthetic data can effectively bridge the gap in model alignment and pre-training. Phi-4 \cite{Phi-4} relies heavily on synthetic data for reasoning capabilities.
In the context of multilingual pre-training, \citet{Wang2025} and \citet{Ji2025} show that machine-translated data from high-resource languages can significantly enhance multilingual pre-training, effectively transferring missing knowledge to low-resource languages.

\subsection{Continued Pre-training}
The release of powerful open-weight models has broadened access to state-of-the-art language technology. Recent families such as Llama 3.1~\cite{llama3}, Qwen 3~\cite{Qwen3}, and Gemma 3~\cite{Gemini3} provide strong foundations for downstream adaptation. For languages and domains that are underrepresented during initial pre-training, CPT remains a primary adaptation approach~\cite{Gururangan2020}. CPT has been used to build some of the strongest BERT-based models for African languages, including the AfroXLMR series~\citep{alabi-etal-2022-adapting,SIB-200,SSA-COMET}.

In the LLM setting, recent work has studied more efficient CPT strategies, such as learning-rate re-warming~\citep{Gupta2023} and replay buffers~\citep{Ibrahim2024}, to reduce catastrophic forgetting. Building on these ideas, \citet{uemura-etal-2024-afriinstruct} and \citet{Lugha-Llama} adapt open LLMs to African languages, releasing AfriInstruct and Lugha-Llama and showing that CPT can yield substantial gains without training from scratch.

Our work builds on CPT and explores new CPT data mixtures to develop \textbf{AfriqueLLM}, a suite of models adapted to the linguistic and cultural diversity of Africa.

\section{AfriqueLLM: Data \& Training Recipe}
\subsection{Dataset Curation}
High-quality and diverse training data is essential for effective language modeling. To mitigate data scarcity for African languages, we curate a 26B-token corpus designed for continued pre-training (CPT).\footnote{All token counts reported in this paper are computed using the Gemma 3 tokenizer.} Our corpus combines monolingual text with code, mathematics, and domain-specific synthetic data to better cover the knowledge and skill distributions needed for downstream tasks. We describe the resulting data pipeline below.

\paragraph{African Monolingual Data.} 
We collect text for the 20 most resource-rich African languages by combining three complementary sources (Table~\ref{tab:unimax_dist}): FineWeb2~\cite{FineWeb2}, WURA~\cite{WURA}, and MADLAD-400~\cite{MADLAD-400}. FineWeb2 provides the backbone of our corpus due to its scale and strong filtering. We add document-level data from WURA to increase contextual diversity and longer-range coherence, and we use MADLAD-400 to improve coverage for the lower-resource languages in our set. To mitigate catastrophic forgetting during CPT, we include four high-resource languages, English, French, Portuguese, and Arabic, capped at 1B tokens per language, following \citet{WURA}. Detailed corpus statistics appear in Table~\ref{tab:dataset-detail} in Appendix~\ref{app:data_details}.

\paragraph{Sampling Strategy.} 

African-language corpora are highly imbalanced, which can cause high-resource languages to dominate training. To mitigate this, we use UniMax sampling~\cite{unimax}, which caps each high-resource language at approximately 1B tokens and upsamples lower-resource languages for up to five epochs. This produces a more balanced sampling distribution and increases coverage of underrepresented languages (see the \textit{UniMax} column in~\autoref{tab:unimax_dist}).

\begin{table}[t]
\centering
\small
\begin{adjustbox}{max width=\columnwidth}
\begin{tabular}{llcccc}
\toprule
\textbf{Language} & \textbf{Code} & \textbf{Raw} & \textbf{Ep.} & \textbf{UniMax} & \textbf{\underline{S}yn.} \\
\midrule
\multicolumn{6}{l}{\textit{High-Resource (Non-African)}} \\
English & eng\_Latn & $>$1.00B & 0 & 1.07B & 16M \\
French & fra\_Latn & $>$1.00B & 0 & 1.07B & -- \\
Portuguese & por\_Latn & $>$1.00B & 0 & 1.07B & -- \\
Arabic & arb\_Arab & $>$1.00B & 0 & 1.07B & -- \\
\midrule
\multicolumn{6}{l}{\textit{African Languages}} \\
Afrikaans & afr\_Latn & 5.30B & 0 & 1.07B & 12M \\
Swahili & swh\_Latn & 2.92B & 0 & 1.07B & 13M \\
Moroccan Ar. & ary\_Arab & 3.29B & 0 & 1.07B & -- \\
Somali & som\_Latn & 1.78B & 0 & 1.07B & 14M \\
Amharic & amh\_Ethi & 989M & 1 & 1.07B & 24M \\
Egyptian Ar. & arz\_Arab & 953M & 1 & 1.07B & -- \\
Hausa & hau\_Latn & 500M & 2 & 1.07B & 13M \\
Kinyarwanda & kin\_Latn & 481M & 2 & 1.07B & 13M \\
Zulu & zul\_Latn & 350M & 3 & 1.07B & 12M \\
Igbo & ibo\_Latn & 318M & 3 & 1.07B & 13M \\
Plateau Malagasy & plt\_Latn & 310M & 3 & 1.07B & 14M \\
Xhosa & xho\_Latn & 268M & 3 & 1.07B & 15M \\
Shona & sna\_Latn & 263M & 4 & 1.05B & 11M \\
Yoruba & yor\_Latn & 258M & 4 & 1.03B & 17M \\
Nyanja & nya\_Latn & 230M & 4 & 921M & 11M \\
Southern Sotho & sot\_Latn & 203M & 4 & 813M & 14M \\
Tigrinya & tir\_Ethi & 142M & 4 & 569M & 76M \\
Tunisian Ar. & aeb\_Arab & 137M & 4 & 547M & -- \\
Oromo & gaz\_Latn & 93M & 4 & 372M & 22M \\
Tswana & tsn\_Latn & 92M & 4 & 368M & 16M \\
\multicolumn{4}{l}{\textit{subtotal}} & \textit{22.8B} & \textit{324M} \\ \midrule
\multicolumn{4}{l}{\textbf{CornStack-Python \cite{CornStack} (\underline{C}ode)}} & \multicolumn{2}{r}{\textit{967M}} \\
\multicolumn{4}{l}{\textbf{FineMath \cite{FineMath} (\underline{M}ath)}} & \multicolumn{2}{r}{\textit{1.07B}} \\
\multicolumn{4}{l}{\textbf{NLLB-OPUS \cite{NLLB} (\underline{P}arallel)}} & \multicolumn{2}{r}{\textit{456M}} \\
\midrule
\multicolumn{4}{l}{\textbf{Total Tokens} --- CM} & \multicolumn{2}{r}{\textbf{24.9B}} \\
\multicolumn{4}{l}{\hspace{1.75cm} --- CMS} & \multicolumn{2}{r}{\textbf{25.2B}} \\
\multicolumn{4}{l}{\hspace{1.75cm} --- CMSP} & \multicolumn{2}{r}{\textbf{25.6B}} \\
\bottomrule
\end{tabular}
\end{adjustbox}
% \vspace{-1em}
\caption{\textbf{Token distribution for the 24 languages pre-trained}. High-resource languages are capped at 1B tokens. Syn. denotes synthetic data.}
\label{tab:unimax_dist}
% \vspace{-2em}
\end{table}

\paragraph{Code (C) and Mathematics (M)} 
Reasoning and logical abilities are often weaker in models adapted to low-resource languages. To strengthen these skills, we incorporate approximately 1B tokens of Python code from CornStack~\cite{CornStack} and approximately 1B tokens of educational mathematics content from FineMath-4+~\cite{FineMath}. We also hypothesize that such structured data acts as a cognitive anchor during CPT. Maintaining a substantial fraction of code and math may help preserve internal consistency and reduce the loss of previously acquired capabilities that can occur when adaptation data is dominated by noisy monolingual web text~\cite{Qwen3,smollm3,SmolLM2}.

\paragraph{Synthetic Data (S)} 
We enrich our training corpus with 324M tokens of machine-translated content drawn from diverse web domains and mathematical reasoning questions to increase topical coverage. Following the domain-centric curation framework of~\citet{OrganizeWeb}, we select 10 domains from Web Organizer~\cite{OrganizeWeb}, which span 20 topics. This design serves two goals. First, it introduces high-quality lexical and conceptual coverage for domains that are sparse in many African-language corpora. Second, it functions as a form of distributional replay buffer~\cite{Gupta2023}: translating high-quality English sources into the target languages helps preserve broad, general-purpose knowledge and stabilizes continued pre-training by keeping the training distribution closer to that of high-resource pre-training.

We use GPT-4.1 for translation due to its strong performance on AfroBench. We translate the selected documents into 17 African languages, excluding Arabic dialects because they are already well represented in our corpus. The resulting translated dataset spans the 10 domains of Food and Dining, Health, History, Industrial, Politics, Science and Technology, Software Development, Travel, Education and Jobs, and Entertainment. In addition, we translate mathematical reasoning questions, thinking traces and solutions from OpenMathReasoning~\cite{OpenMathReasoning} (the \texttt{cot} split) and include them as an eleventh domain.

\paragraph{Translation Data (P)} 
To refine cross-lingual alignment, we explored the integration of parallel data from the NLLB project~\cite{NLLB}. Although we initially collected 1B bilingual pairs, quality control was paramount. We applied a rigorous filtering threshold of 0.7 using SSA-COMET~\cite{SSA-COMET}—a regression model for machine translation (MT) quality estimation (QE) specifically optimized for African languages. This process yielded a high-quality subset of 4M samples (approx. 456M tokens), ensuring that only the most reliable translation pairs contributed to the model's multilingual capabilities.

Overall, our curation process yields several dataset mixtures, like \textsc{CMS} and \textsc{CMSP}, totaling 25.2B and 25.6B tokens respectively, with the detailed per-language distribution across all 24 training languages presented in Table~\ref{tab:unimax_dist} and further elaborated in Appendix \ref{app:detailed_results}.

\subsection{Training Setup}
Experiments were conducted using the LLaMA-Factory \cite{llamafactory} framework on a high-performance cluster (up to 16 nodes, 64 NVIDIA H100 GPUs). We maximized training throughput and memory efficiency by employing sequence packing, DeepSpeed ZeRO-1/ZeRO-2 \cite{deepspeed}, Flash Attention 3 \cite{flashattn3}, and Liger Kernel \cite{ligerkernel}.

\paragraph{Hyperparameter Tuning}
Following the continual pre-training strategies of \citet{gupta2023continual} and \citet{smollm3}, we performed an extensive ablation study to tailor hyperparameters for the African language context. Our search yielded three key insights based on the \texttt{gemma-3-4b/12b-pt}:
% \vspace{-0.6em}
\begin{enumerate}[leftmargin=*]\setlength\itemsep{0pt}\setlength\parskip{0pt}\setlength\leftmargin{0pt}
    \item \textbf{Learning Rate:} A sweep from $1\text{e-}6$ to $2\text{e-}4$ revealed that $5\text{e-}5$ optimally balances the retention of prior knowledge with the acquisition of new linguistic features.
    \item \textbf{Context Length:} Evaluating window sizes of 4k, 16k, and 32k tokens, we found that the 16k sequence length provided the best performance on reasoning tasks such as AfriMGSM.
    \item \textbf{Learning Rate Scheduler:} We fine-tuned the cosine scheduler, setting a minimum learning rate ratio of 0.01 and a warmup ratio of 0.001 to ensure training stability.
\end{enumerate}
% \vspace{-0.6em}
We maintained a global batch size of 4M tokens across all runs, dynamically adjusting gradient accumulation steps to accommodate varying hardware configurations. Full configuration details and grid search results are available in Appendix \ref{app:training_config}.

\section{Evaluation Setting}
We use a comprehensive evaluation suite to assess model performance across Africa's diverse linguistic landscape. Our primary benchmark is AfroBench \cite{AfroBench}, which covers 64 languages across 15 tasks.

\paragraph{AfroBench-Lite}
To facilitate efficient yet comprehensive evaluation, we focus on the AfroBench-Lite subset, which selects 7 representative tasks/datasets covering key capabilities: AfriMGSM (Math), AfriMMLU (Knowledge), AfriXNLI (natural language inference) \cite{Adelani2024IrokoBenchAN}, Belebele (Reading Comprehension) \cite{bandarkar-etal-2024-belebele}, Flores (Translation) \cite{goyal-etal-2022-flores}, Injongo (Intent Classification) \cite{Yu2025INJONGOAM}, and SIB (topic classification) \cite{adelani-etal-2024-sib}. While the original AfroBench-Lite evaluated on only 14 languages, we expanded the coverage of our evaluation to all African languages covered in each dataset/task.

\paragraph{Metrics}
We strictly adhere to the \texttt{lm-eval}~\cite{eval-harness} tasks established by AfroBench to ensure comparability. Note that as our models are pre-trained checkpoints without any instruction tuning, we report few-shot (5-shots) results for all tasks except AfriMGSM (where the default setting is 8-shots). For translation tasks (Flores), we utilize SSA-COMET \cite{SSA-COMET} rather than lexical overlap metrics like ChrF++~\citep{popovic-2017-chrf} from the official AfroBench since recent studies indicate that SSA-COMET correlates significantly better with human judgment for African languages, offering a more accurate assessment of semantic quality \cite{SSA-COMET}. All evaluations utilize the Hugging Face or vLLM backend \cite{vllm} with ``do\_sample=False''.

\paragraph{Baseline Models}
To evaluate the effectiveness of our data mixture and scaling laws, we selected several state-of-the-art open-weight models as baselines: the Google Gemma 3 series \cite{gemma3_2025}, Meta Llama 3.1 series \cite{llama3}, and Alibaba Qwen 3 series \cite{yang2025qwen3}. Gemma 3 is renowned for its extensive multilingual support, while Llama 3.1 represents a highly optimized predecessor in the open-source landscape. 
We included the Qwen 3 series due to its strong performance in mathematical reasoning, despite its limited native support for African languages.\!\footnote{\href{https://qwenlm.github.io/blog/qwen3/\#key-features}{Qwen 3 supported languages}} 
Our experimental pipeline first validates the data mixture using Gemma 3 (4B and 12B) and subsequently scales these findings to Llama 3.1 (8B) and Qwen 3 (4B, 8B and 14B) base models.

\section{Experiments Results}
\subsection{Data Mixture Ablation}
\label{sec:data-mixture}
To identify the optimal recipe for African language adaptation, we perform an ablation study on Gemma 3 (4B and 12B), evaluating four benchmarks: Flores (MT), AfriXNLI (NLI), AfriMGSM (Math), and AfriMMLU (QA). Results are shown in Table~\ref{tab:data-composition}.\footnote{For the data mixture ablation study, we use the HuggingFace backend with \texttt{lm-eval} for accuracy, while all other benchmarks use vLLM to reduce computation cost. As a result, relative trends are consistent, but absolute scores may differ between Table~\ref{tab:data-composition} and Table~\ref{tab:task_results}.}

\begin{table}[th]
\centering
\footnotesize
\resizebox{\linewidth}{!}{%
\begin{tabular}{l|cccc}
\toprule
\textbf{Model} & \textbf{Flores} & \textbf{AfriMGSM} & \textbf{AfriMMLU} & \textbf{AfriXNLI} \\
\midrule
\multicolumn{5}{c}{\textit{Baseline Models}} \\
\midrule
NLLB-200-1.3B & 61.27 & -- & -- & -- \\
NLLB-200-3.3B & 62.42 & -- & -- & -- \\
NLLB-MoE-54B & 65.72 & -- & -- & -- \\
\midrule
\multicolumn{5}{c}{\textit{Gemma 3 4B Variants}} \\
\midrule
Gemma 3 4B PT & 35.99 & 9.25 & 33.57 & 34.77 \\
Gemma 3 4B IT & 31.86 & 14.29 & 34.44 & 33.19 \\
\midrule
+ Monolingual & 62.72 {\scriptsize\textcolor{green!50!black}{$\uparrow$}} & 10.68 {\scriptsize\textcolor{green!50!black}{$\uparrow$}} & 35.41 {\scriptsize\textcolor{green!50!black}{$\uparrow$}} & \textbf{40.76} {\scriptsize\textcolor{green!50!black}{$\uparrow$}} \\
+ CM & 62.30 {\scriptsize\textcolor{green!50!black}{$\uparrow$}} & 14.68 {\scriptsize\textcolor{green!50!black}{$\uparrow$}} & 36.08 {\scriptsize\textcolor{green!50!black}{$\uparrow$}} & 40.19 {\scriptsize\textcolor{green!50!black}{$\uparrow$}} \\
+ CMP & 63.21 {\scriptsize\textcolor{green!50!black}{$\uparrow$}} & 14.29 {\scriptsize\textcolor{green!50!black}{$\uparrow$}} & 35.20 {\scriptsize\textcolor{green!50!black}{$\uparrow$}} & 40.10 {\scriptsize\textcolor{green!50!black}{$\uparrow$}} \\
+ CMS & 63.17 {\scriptsize\textcolor{green!50!black}{$\uparrow$}} & \textbf{14.81} {\scriptsize\textcolor{green!50!black}{$\uparrow$}} & 35.86 {\scriptsize\textcolor{green!50!black}{$\uparrow$}} & 39.93 {\scriptsize\textcolor{green!50!black}{$\uparrow$}} \\
+ CMSP & \textbf{63.34} {\scriptsize\textcolor{green!50!black}{$\uparrow$}} & 13.35 {\scriptsize\textcolor{green!50!black}{$\uparrow$}} & \textbf{36.72} {\scriptsize\textcolor{green!50!black}{$\uparrow$}} & 40.44 {\scriptsize\textcolor{green!50!black}{$\uparrow$}} \\
\midrule
\multicolumn{5}{c}{\textit{Gemma 3 12B Variants}} \\
\midrule
Gemma 3 12B PT & 52.53 & 24.10 & 48.21 & 39.81 \\
Gemma 3 12B IT & 47.81 & 36.50 & 46.84 & 40.16 \\
\midrule
+ Monolingual & 65.78 {\scriptsize\textcolor{green!50!black}{$\uparrow$}} & 23.78 {\scriptsize\textcolor{red}{$\downarrow$}} & 46.72 {\scriptsize\textcolor{red}{$\downarrow$}} & \textbf{45.19} {\scriptsize\textcolor{green!50!black}{$\uparrow$}} \\
+ CMP & 65.86 {\scriptsize\textcolor{green!50!black}{$\uparrow$}} & 27.82 {\scriptsize\textcolor{green!50!black}{$\uparrow$}} & \textbf{48.49} {\scriptsize\textcolor{green!50!black}{$\uparrow$}} & 42.45 {\scriptsize\textcolor{green!50!black}{$\uparrow$}} \\
\rowcolor[HTML]{E8F5E9}+ CMS & \textbf{66.23} {\scriptsize\textcolor{green!50!black}{$\uparrow$}} & \textbf{30.87} {\scriptsize\textcolor{green!50!black}{$\uparrow$}} & 48.46 {\scriptsize\textcolor{green!50!black}{$\uparrow$}} & 44.57 {\scriptsize\textcolor{green!50!black}{$\uparrow$}} \\
+ CMSP & 65.83 {\scriptsize\textcolor{green!50!black}{$\uparrow$}} & 29.61 {\scriptsize\textcolor{green!50!black}{$\uparrow$}} & 48.32 {\scriptsize\textcolor{green!50!black}{$\uparrow$}} & 43.26 {\scriptsize\textcolor{green!50!black}{$\uparrow$}} \\
\bottomrule
\end{tabular}
}
\vspace{1mm}
\caption{\textbf{Ablation of CPT Data mixture}. We report result on African languages covered in CPT. %Covered languages are provided at Appendix Table~\ref{tab:benchmarks-langs}. 
C = Code, M = Math, S = Synthetic, P = Parallel. Best results among adapted models are in \textbf{bold}, and our final configurations are highlighted in \colorbox[HTML]{E8F5E9}{green}. Compare to base model: improvement with \textcolor{green!50!black}{$\uparrow$} and degradation with \textcolor{red}{$\downarrow$}.}
\label{tab:data-composition}
\vspace{-2em}
\end{table}

\begin{table*}[tbp]
\centering
\small
\resizebox{\textwidth}{!}{
\begin{tabular}{lcccccccccc}
\toprule
\textbf{Model} & \textbf{AfriMGSM} & \textbf{AfriMMLU} & \textbf{AfriXNLI} & \textbf{Belebele} & \textbf{Flores} & \textbf{Injongo} & \textbf{SIB-200} & \textbf{Overall} & \textbf{$\Delta$} & \textbf{$\Delta$ \%} \\
\midrule
\textit{African Languages Adapted} \\
Lugha-Llama-8B-wura & 9.46 & 37.00 & 39.24 & 47.86 & 49.90 & 62.30 & 75.81 & 45.94 & - & - \\
\midrule
\textit{Base Models} \\
Llama 3.1 8B & 8.14 & 32.27 & 37.90 & 40.95 & 26.69 & 41.37 & 59.99 & 35.33 & - & - \\
Gemma 3 4B & 10.24 & 33.89 & 37.76 & 45.79 & 35.36 & 55.52 & 63.59 & 40.31 & - & - \\
Gemma 3 12B & 25.21 & 48.76 & 44.01 & 68.84 & 44.09 & 73.53 & 79.17 & 54.80 & - & - \\
Qwen 3 4B & 8.26 & 33.84 & 37.12 & 41.50 & 20.16 & 21.69 & 57.88 & 31.49 & - & - \\
Qwen 3 8B & 11.22 & 36.56 & 38.24 & 44.63 & 21.13 & 29.47 & 53.06 & 33.47 & - & - \\
Qwen 3 14B & 16.60 & 39.66 & 43.22 & 50.74 & 23.75 & 41.80 & 66.29 & 40.29 & - & - \\
\midrule
\textit{Afrique Models (Ours)} \\
AfriqueLlama-8B & 17.51 & 36.57 & 37.39 & 50.51 & 63.60 & 71.17 & 69.14 & 49.41 & \textcolor{blue}{+14.1} & \textcolor{blue}{+39.9\%} \\
AfriqueGemma-4B & 14.86 & 36.73 & 39.62 & 50.52 & 54.95 & 69.28 & 69.21 & 47.88 & \textcolor{blue}{+7.6} & \textcolor{blue}{+18.8\%} \\
AfriqueGemma-12B & 32.14 & 49.47 & 44.60 & 68.65 & \textbf{65.04} & 76.79 & 75.08 & 58.82 & \textcolor{blue}{+4.0} & \textcolor{blue}{+7.3\%} \\
AfriqueQwen-4B & 33.09 & 43.04 & 44.88 & 63.62 & 59.82 & 65.34 & 74.77 & 54.94 & \textcolor{purple}{+23.4} & \textcolor{purple}{+74.4\%} \\
AfriqueQwen-8B & \underline{39.68} & 46.91 & 45.99 & 68.46 & 62.18 & 73.36 & 77.00 & 59.08 & \textcolor{purple}{\textbf{+25.6}} & \textcolor{purple}{\textbf{+76.5\%}} \\
AfriqueQwen-14B & \textbf{45.01} & \underline{52.22} & \textbf{49.01} & \underline{74.63} & \underline{63.77} & \underline{77.80} & \underline{82.63} & \textbf{63.58} & \textcolor{purple}{\underline{+23.3}} & \textcolor{purple}{\underline{+57.8\%}} \\
\midrule
Gemma 3 27B & 35.37 & \textbf{55.47} & \underline{46.85} & \textbf{74.81} & 48.41 & \textbf{79.70} & \textbf{84.34} & \underline{60.71} & - & - \\

\bottomrule
\end{tabular}
}
% \vspace{-2mm}
\caption{\textbf{Task-level performance comparison between Base models and our Continued Pre-Trained (CPT) models over only CPT languages}. Best results are \textbf{bolded}, second-best are \underline{underlined}. $\Delta$Abs and $\Delta$Rel show absolute and relative improvements over base models, with \textcolor{purple}{purple} highlighting Qwen's superior gains.}
\label{tab:task_results}
% \vspace{-1em}
\end{table*}

\paragraph{The Monolingual Trade-off}
Adding only monolingual data (22B tokens) yields substantial gains on non-reasoning tasks, with MT (Flores) and NLI (AfriXNLI) improving by over 10\% relative to the base model. However, for the 12B model, challenging reasoning datasets (\textsc{AfriMGSM} and \textsc{AfriMMLU}) decline slightly (e.g., 24.1 $\rightarrow$ 23.8 on AfriMGSM and 48.2 $\rightarrow$ 46.7 on AfriMMLU). We attribute this to catastrophic forgetting of reasoning priors when exposed to large volumes of raw web text that are less heterogeneous for low-resource languages.  %without logical reinforcement.

\paragraph{Performance Recovery via Code and Math}
Integrating additional 2B tokens of Code and Math (\textsc{CM}) reverses this trend. For the 4B model, \textsc{CM} improves performance across all tasks compared to monolingual data only. For the 12B model, all configurations that include code and math (CMP, CMS, CMSP) similarly recover reasoning performance, demonstrating the importance of adding datasets with structured reasoning such as Code. This finding aligns with prior work showing that \textsc{CM} enhances generalization to other tasks~\citep{allal2025smollm,aryabumi2025to}.

\paragraph{Data Quality vs. Scale}
At 12B scale, we observe a divergence regarding parallel data (\textsc{P}). While NLLB parallel data (\textsc{CMP}) provides marginal gains for the 4B model, it becomes detrimental for the 12B model compared to \textsc{CMS}. Specifically, \textsc{CMS} (Monolingual + Code/Math + Synthetic) achieves the highest scores on MGSM (30.9) and Flores (66.2), whereas adding parallel data (\textsc{CMSP}) causes performance reduction.

Drawing from mid-training recipes in \citet{Hunyuan-MT,smollm3}, we hypothesize that larger models are more sensitive to data quality: noisy parallel corpora like NLLB, even when filtered, benefit smaller models but harm larger ones. Accordingly, we adopt \textbf{\textsc{CMS}} as our primary recipe.

\begin{tcolorbox}[
  enhanced,
  colback=green!5!white,
  colframe=green!60!black,
  arc=4mm,
  boxrule=0.8pt,
  left=3mm,
  right=3mm,
  top=2mm,
  bottom=2mm,
  fonttitle=\small\sffamily\bfseries,
  title={\textcolor{white}{$\triangleright$ Takeaway 1: The Quality-Scale Sensitivity}},
  coltitle=white,
  colbacktitle=green!60!black,
  attach boxed title to top left={yshift=-2mm, xshift=3mm},
  boxed title style={arc=2mm, boxrule=0pt}
]
\textit{Leveraging synthetic data and datasets with structured reasoning such as Math \& Code improves CPT generalization}. For larger models (12B+), high-quality synthetic data is a more effective bridge than noisy parallel corpora.
\end{tcolorbox}

\subsection{Impact of Model Selection and Scaling}

Table~\ref{tab:task_results} shows the result of leveraging the \textsc{CMS} recipe across various model architectures and model sizes. 
\paragraph{``Zero-to-Hero'' Effect in Qwen 3}
The most striking finding is the performance jump in the Qwen 3 series, with relative improvements of 74.4\% (4B), 76.5\% (8B) and 57.8\% (14B) over their respective base models, while the Gemma 3 series achieved 18.8\% (4B) and 7.3\% (12B) relative improvement. We term this the ``Zero-to-Hero'' effect. Despite minimal official support for African languages and the weakest baseline performance (Qwen 3 8B avg.: 33.47), AfriqueQwen exhibits the highest relative gains, outperforming similarly-sized AfriqueGemma variants on all tasks except translation (Flores), where Gemma's native multilingual pre-training provides an expected advantage. We extend this analysis to the broader 4B Qwen family in Section~\ref{sec:4b-qwen}.
And even notably, AfriqueQwen-14B (63.58) outperforms Gemma 3 27B (60.71) by +2.87 points overall, with significant advantages on AfriMGSM (+9.64) and Flores (+15.36), despite being less than half the size.

These results suggest that Qwen 3 models largely preserve their High-Resource Languages (HRLs) performance when adapted to Low-Resource Languages (LRLs) via CPT. Consistent with the Qwen 3 technical report, Qwen 3 14B outperforms Gemma 3 12B on HRLs. We hypothesize that Qwen 3 benefits from stronger latent fast adaptation capabilities that are more effectively unlocked through CPT,~\footnote{Probably because it was pre-trained on 119 languages} highlighting that a strong HRL base model priors are more critical for cross-lingual adaptation than prior language familiarity.

\begin{figure*}[t]
    \centering
    \includegraphics[width=0.8\linewidth]{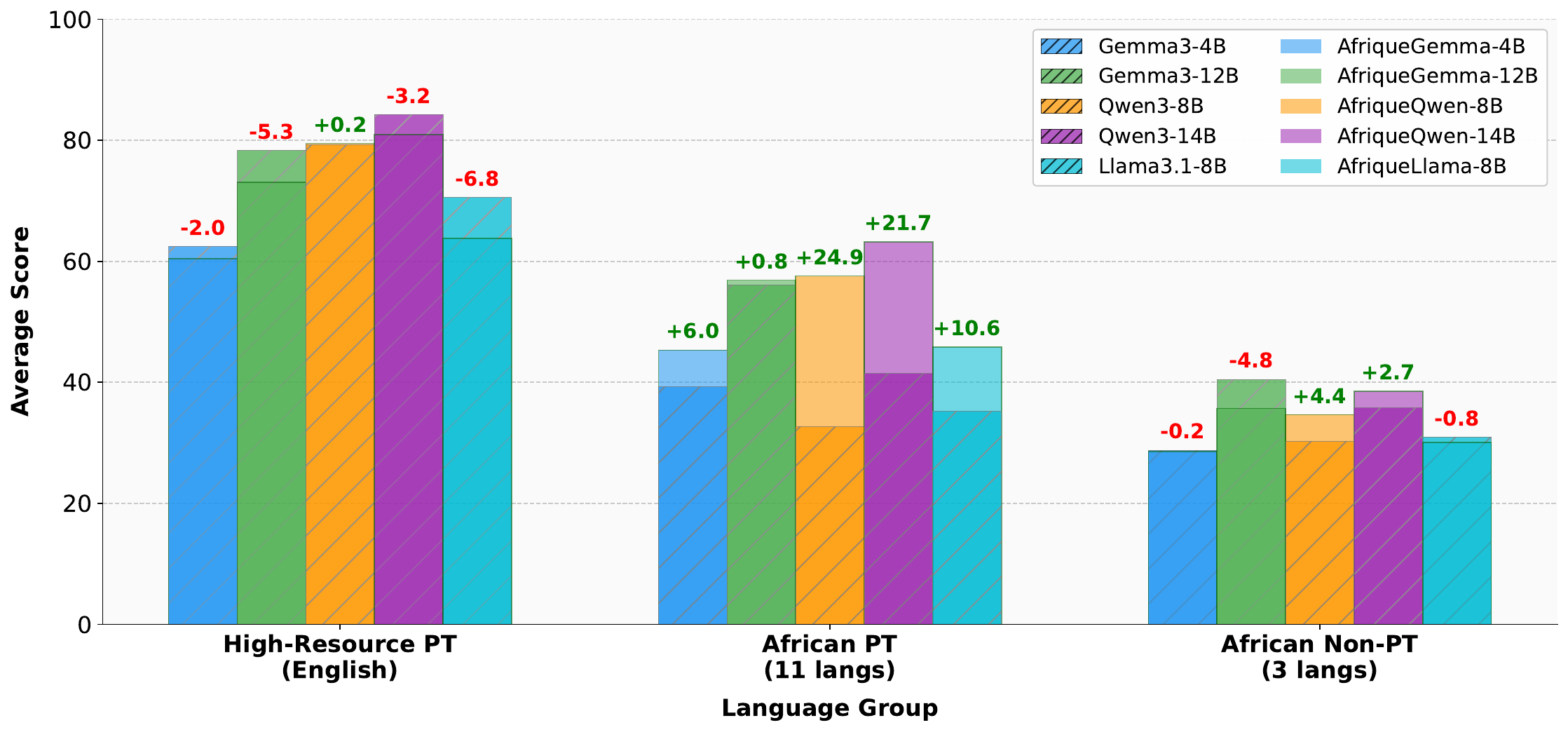}
    \vspace{-1em}
    \caption{\textbf{Performance comparison across language groups: High-Resource PT (English), African PT, and African Non-PT}. We report the average score across all benchmarks excluding Flores. Hatched bars represent base models, while solid bars represent their Afrique-adapted counterparts. Values above the bars indicate the absolute improvement ($\Delta$) after adaptation.}
    \vspace{-0.5em}
    \label{fig:language_group_scores}
\end{figure*}

\begin{table*}[t]
\centering
\resizebox{\textwidth}{!}{%
\begin{tabular}{l|ccc|ccc|ccc|ccc|ccc}
\toprule
& \multicolumn{3}{c|}{\textbf{Gemma3-4B}} & \multicolumn{3}{c|}{\textbf{Gemma3-12B}} & \multicolumn{3}{c|}{\textbf{Qwen3-8B}} & \multicolumn{3}{c|}{\textbf{Qwen3-14B}} & \multicolumn{3}{c}{\textbf{Llama3.1-8B}} \\
\textbf{Language} & Base & Afrique & $\Delta$\% & Base & Afrique & $\Delta$\% & Base & Afrique & $\Delta$\% & Base & Afrique & $\Delta$\% & Base & Afrique & $\Delta$\% \\
\midrule
English & 59.1 & 56.6 & \textcolor{red}{-4.2\%} & 76.9 & 71.2 & \textcolor{red}{-7.4\%} & 78.1 & 78.7 & \textcolor{green!50!black}{+0.8\%} & 83.4 & 79.9 & \textcolor{red}{-4.1\%} & 68.1 & 60.6 & \textcolor{red}{-11.0\%} \\
French & 49.6 & 45.5 & \textcolor{red}{-8.3\%} & 66.5 & 64.0 & \textcolor{red}{-3.8\%} & 73.9 & 71.0 & \textcolor{red}{-4.0\%} & 76.7 & 74.1 & \textcolor{red}{-3.5\%} & 55.0 & 49.8 & \textcolor{red}{-9.4\%} \\
\midrule
\textbf{Avg.} & 54.3 & 51.1 & \textcolor{red}{-6.2\%} & 71.7 & 67.6 & \textcolor{red}{-5.6\%} & 76.0 & 74.8 & \textcolor{red}{-1.6\%} & 80.0 & 77.0 & \textcolor{red}{-3.8\%} & 61.5 & 55.2 & \textcolor{red}{-10.2\%} \\
\bottomrule
\end{tabular}%
}
\vspace{-2mm}
\caption{\textbf{High-Resource Language (HRL) performance comparison between base models and Afrique-adapted models}. Red values indicate performance drops, highlighting the trade-off when adapting models for African languages. $\Delta$\% denotes relative difference.}
    \vspace{-1em}
\label{tab:hrl-drop}
\end{table*}

% Tabel 5: Language-wise all task improved
\begin{table*}[h]
\centering
\resizebox{0.9\linewidth}{!}{%
\begin{tabular}{l|ccccccccccc|c}
\toprule
\textbf{Model} & \textbf{amh} & \textbf{hau} & \textbf{ibo} & \textbf{kin} & \textbf{orm} & \textbf{sna} & \textbf{sot} & \textbf{swa} & \textbf{xho} & \textbf{yor} & \textbf{zul} & \textbf{Avg.} \\ \midrule
Llama 3.1 8B & 29.0 & 41.0 & 37.5 & 29.2 & 25.8 & 28.7 & 28.0 & 48.7 & 28.7 & 29.5 & 28.9 & 32.3 \\
\rowcolor[HTML]{E8F5E9} AfriqueLlama-8B & 47.9 & 50.3 & 47.2 & 47.5 & 45.3 & 50.1 & 48.0 & 55.0 & 48.6 & 47.4 & 47.0 & 48.6 \\
\multicolumn{1}{r|}{\textit{$\Delta$}} & \textcolor{green!50!black}{+18.9} & \textcolor{green!50!black}{+9.3} & \textcolor{green!50!black}{+9.8} & \textcolor{green!50!black}{+18.3} & \textcolor{green!50!black}{+19.5} & \textcolor{green!50!black}{+21.5} & \textcolor{green!50!black}{+20.1} & \textcolor{green!50!black}{+6.3} & \textcolor{green!50!black}{+19.9} & \textcolor{green!50!black}{+17.9} & \textcolor{green!50!black}{+18.1} & \textcolor{green!50!black}{+16.3} \\
\midrule
Gemma 3 4B & 43.3 & 42.6 & 37.3 & 36.3 & 26.4 & 38.2 & 33.2 & 52.4 & 38.0 & 26.6 & 38.8 & 37.6 \\
\rowcolor[HTML]{E8F5E9} AfriqueGemma-4B & 48.7 & 50.0 & 46.4 & 46.9 & 42.5 & 49.8 & 43.0 & 54.1 & 48.4 & 43.8 & 46.5 & 47.3 \\
\multicolumn{1}{r|}{\textit{$\Delta$}} & \textcolor{green!50!black}{+5.4} & \textcolor{green!50!black}{+7.4} & \textcolor{green!50!black}{+9.1} & \textcolor{green!50!black}{+10.6} & \textcolor{green!50!black}{+16.1} & \textcolor{green!50!black}{+11.6} & \textcolor{green!50!black}{+9.7} & \textcolor{green!50!black}{+1.7} & \textcolor{green!50!black}{+10.4} & \textcolor{green!50!black}{+17.2} & \textcolor{green!50!black}{+7.7} & \textcolor{green!50!black}{+9.7} \\ \midrule
Gemma 3 12B & 59.9 & 57.8 & 52.8 & 52.7 & 41.4 & 56.7 & 51.1 & 66.8 & 53.2 & 43.6 & 53.7 & 53.6 \\
\rowcolor[HTML]{E8F5E9} AfriqueGemma-12B & 60.8 & 60.4 & 55.8 & 56.1 & 54.5 & 59.8 & 58.2 & 66.3 & 58.0 & 54.6 & 57.5 & 58.3 \\
\multicolumn{1}{r|}{\textit{$\Delta$}} & \textcolor{green!50!black}{+0.9} & \textcolor{green!50!black}{+2.6} & \textcolor{green!50!black}{+3.0} & \textcolor{green!50!black}{+3.4} & \textcolor{green!50!black}{+13.1} & \textcolor{green!50!black}{+3.0} & \textcolor{green!50!black}{+7.0} & \textcolor{red}{-0.5} & \textcolor{green!50!black}{+4.8} & \textcolor{green!50!black}{+11.0} & \textcolor{green!50!black}{+3.8} & \textcolor{green!50!black}{+4.7} \\ \midrule
Qwen 3 8B & 34.6 & 24.6 & 25.6 & 24.9 & 28.4 & 27.4 & 28.3 & 47.3 & 27.2 & 24.8 & 25.8 & 29.0 \\
\rowcolor[HTML]{E8F5E9} AfriqueQwen-8B & 61.0 & 61.7 & 54.8 & 56.8 & 55.1 & 59.6 & 57.3 & 68.2 & 57.2 & 55.0 & 56.4 & 58.5 \\
\multicolumn{1}{r|}{\textit{$\Delta$}} & \textbf{\textcolor{green!50!black}{+26.3}} & \textbf{\textcolor{green!50!black}{+37.1}} & \textbf{\textcolor{green!50!black}{+29.2}} & \textbf{\textcolor{green!50!black}{+31.9}} & \textbf{\textcolor{green!50!black}{+26.8}} & \textbf{\textcolor{green!50!black}{+32.2}} & \textbf{\textcolor{green!50!black}{+29.0}} & \textbf{\textcolor{green!50!black}{+20.8}} & \textbf{\textcolor{green!50!black}{+30.0}} & \textbf{\textcolor{green!50!black}{+30.2}} & \textbf{\textcolor{green!50!black}{+30.6}} & \textbf{\textcolor{green!50!black}{+29.5}} \\ \midrule
Qwen 3 14B & 42.0 & 32.2 & 32.9 & 31.0 & 35.4 & 33.3 & 35.8 & 58.7 & 36.7 & 33.7 & 35.1 & 37.0 \\
\rowcolor[HTML]{E8F5E9} AfriqueQwen-14B & 64.7 & 66.1 & 61.0 & 61.0 & 62.0 & 64.5 & 62.4 & 73.0 & 61.8 & 61.2 & 61.5 & 63.6 \\
\multicolumn{1}{r|}{\textit{$\Delta$}} & {\ul \textcolor{green!50!black}{+22.7}} & {\ul \textcolor{green!50!black}{+34.0}} & {\ul \textcolor{green!50!black}{+28.1}} & {\ul \textcolor{green!50!black}{+30.0}} & {\ul \textcolor{green!50!black}{+26.7}} & {\ul \textcolor{green!50!black}{+31.2}} & {\ul \textcolor{green!50!black}{+26.6}} & {\ul \textcolor{green!50!black}{+14.3}} & {\ul \textcolor{green!50!black}{+25.1}} & {\ul \textcolor{green!50!black}{+27.5}} & {\ul \textcolor{green!50!black}{+26.4}} & {\ul \textcolor{green!50!black}{+26.6}} \\
\bottomrule
\end{tabular}%
}
\vspace{-1.5mm}
\caption{\textbf{Language-wise average performance improvement across all benchmarks on CPT covered languages}. Green values indicate gains after Afrique adaptation. \textbf{Bold} and {\ul underline} denote the best and second-best improvements per language.}
\label{tab:cpted-language-improvement}
\end{table*}

\paragraph{Comparison with Other CPT African LLMs}
Compared to \textbf{Lugha-Llama-8B-wura}~\cite{Lugha-Llama}, adapted using only WURA monolingual data~\cite{WURA} on the same Llama 3.1 8B base, AfriqueLlama shows close score to Lugha, and outperforms it in 4 of 7 tasks, particularly reasoning (MGSM: 17.51 vs. 9.46) and translation (Flores: 63.60 vs. 49.90).
\paragraph{Marginal Effects of Model Size}
As expected, relative improvement from CPT decreases with model size (Gemma 4B: +18.8\% vs. 12B: +7.3\%) with the same training data mixture, consistent with scaling laws in prior work~\cite{Ye2024,He2024}. However, even at 14B parameters, Qwen shows substantial gains (+57.8\%), indicating significant headroom for African language adaptation.

\begin{tcolorbox}[
  enhanced,
  colback=blue!5!white,
  colframe=blue!60!black,
  arc=4mm,
  boxrule=0.8pt,
  left=3mm,
  right=3mm,
  top=2mm,
  bottom=2mm,
  fonttitle=\small\sffamily\bfseries,
  title={\textcolor{white}{$\triangleright$ Takeaway 2: Foundation Ability Matters}},
  coltitle=white,
  colbacktitle=blue!60!black,
  attach boxed title to top left={yshift=-2mm, xshift=3mm},
  boxed title style={arc=2mm, boxrule=0pt}
]
\textit{A base model's ``strong ability'' is a more potent starting point for CPT than its ``language coverage.''} Strong foundation ability priors can be effectively mapped to new languages with high-quality data mixture.
\end{tcolorbox}

\subsection{Language-wise Analysis}
Here, we analyze the impact of CPT across three language resource levels (Figure~\ref{fig:language_group_scores}): High-Resource Pre-Trained (HRL-PT) language --- English that is well-represented in base model pre-training; African Pre-Trained (Afr-PT) languages included in our CPT corpus (e.g., Swahili, Amharic); and African Non-Pre-Trained (Afr-NPT) languages absent from both base and CPT training (e.g., Ewe, Lingala).
Figure~\ref{fig:language_group_scores} reveals three key findings: \textit{(1) Targeted gains on Afr-PT languages.} All models show substantial improvements on CPT-covered African languages, with Qwen 3 8B achieving the highest gain (+24.9 points). \textit{(2) Minimal transfer to unseen languages.} Performance on Afr-NPT languages remains largely unchanged for most models, indicating that CPT primarily benefits explicitly covered languages. Interestingly, AfriqueQwens show modest positive transfer (+4.4, +2.7), suggesting that the CPT models leverage cross-lingual transfer from related languages from same family e.g. Lingala could benefit from other Bantu languages (like Swahili \& Kinyarwanda) even when not covered.
\textit{(3) Less catastrophic forgetting for HRLs} While most models exhibit HRL decline, Qwen 3 8B maintains near-parity (+0.2), demonstrating that with a strong HRL base, CPT can enhance low-resource languages without sacrificing too much high-resource performance. This is further supported by AfriqueQwen-14B's +10.6 gain on Afr-PT languages with only -3.2 loss on HRL. Overall, the mixture of prior model capability and CPT data composition allows for balancing improvements in LRLs while controlling degradation in HRLs.

\vspace{-2mm}
\begin{tcolorbox}[
  enhanced,
  colback=orange!5!white,
  colframe=orange!70!black,
  arc=4mm,
  boxrule=0.8pt,
  left=3mm,
  right=3mm,
  top=2mm,
  bottom=2mm,
  fonttitle=\small\sffamily\bfseries,
  title={\textcolor{white}{$\triangleright$ Takeaway 3: Language Transfer Limits}},
  coltitle=white,
  colbacktitle=orange!70!black,
  attach boxed title to top left={yshift=-2mm, xshift=3mm},
  boxed title style={arc=2mm, boxrule=0pt}
]
\textit{``CPT favours seen languages in data mixtures.'' } Including HRLs in the CPT data mixture mitigates catastrophic forgetting on HRLs, but yields limited transfer to unseen languages.
\end{tcolorbox}

\paragraph{HRL degradation across models}
\autoref{tab:hrl-drop} quantifies catastrophic forgetting on English and French. Compared to the massive African language gains (up to +76.5\% in Table~\ref{tab:task_results}), HRL performance drops are contained but noteworthy. Llama 3.1 8B shows the steepest average decline (-10.2\% relative), followed by Gemma 3 models (-5.6\% to -6.2\%). In contrast, the Qwen 3 series exhibits the smallest average HRL degradation (-1.6\% for 8B, -3.8\% for 14B), showing they are slightly better in preventing catastrophic forgetting.

% TODO: start from what we successfully improved
\paragraph{Granular Analysis on Afr-PT Languages}
\autoref{tab:cpted-language-improvement} provides a detailed breakdown across 11 CPT-covered African languages, averaged across tasks. Gains are consistent across all languages and models, with \textbf{many low-resource languages benefiting most}: Oromo (orm) and Yoruba (yor) show the highest deltas (e.g., +16.1 and +17.2 for AfriqueGemma-4B). In contrast, Swahili (swa) shows more modest gains (+1.7 to +20.8). For Qwen 3, improvements are even more astounding: AfriqueQwen-8B exceeds +25 absolute points in 10 of 11 languages, peaking at +37.1 in Hausa. This confirms our hypothesis that previously underrepresented languages benefit most from our data mixture. The more detailed results across all languages and tasks are presented in Appendix~\ref{app:detailed_results}.

\begin{table}[t]
\centering
\small
\resizebox{\linewidth}{!}{%
\begin{tabular}{l|rrrrr|r}
\toprule
\textbf{Model} & \textbf{amh} & \textbf{hau} & \textbf{swa} & \textbf{yor} & \textbf{zul} & \textbf{Avg.} \\
\midrule
\multicolumn{7}{c}{\textit{English $\rightarrow$ African (eng2xx)}} \\
\midrule
\rowcolor{blue!10} Llama 3.1 8B SFT$_{10}$ & 27.6 & 49.7 & 64.1 & \textbf{50.3} & 47.0 & 47.8 \\
Llama 3.1 8B & 10.3 & 19.5 & 28.7 & 16.7 & 14.2 & 17.9 \\
AfriqueLlama-8B & 41.4 & 62.0 & 74.4 & 46.3 & 68.1 & 58.5 \\
AfriqueGemma-12B & \textbf{42.1} & \textbf{64.2} & \textbf{78.1} & 47.0 & \textbf{69.8} & \textbf{60.2} \\
AfriqueQwen-14B & 42.0 & 62.8 & 75.7 & 47.4 & 68.2 & 59.2 \\
\midrule
\multicolumn{7}{c}{\textit{African $\rightarrow$ English (xx2eng)}} \\
\midrule
\rowcolor{blue!10} Llama 3.1 8B SFT$_{10}$ & 63.8 & 61.7 & 74.4 & 68.9 & 71.4 & 68.0 \\
Llama 3.1 8B & 20.0 & 53.9 & 71.2 & 30.7 & 37.0 & 42.6 \\
AfriqueLlama-8B & 44.7 & 58.2 & 66.6 & 53.5 & 63.4 & 57.3 \\
AfriqueGemma-12B & 72.7 & 67.7 & \textbf{80.5} & 68.8 & \textbf{76.6} & 73.3 \\
AfriqueQwen-14B & \textbf{72.8} & \textbf{68.3} & 79.7 & \textbf{70.8} & 76.1 & \textbf{73.5} \\
\bottomrule
\end{tabular}%
}
\vspace{-1.5mm}
\caption{\textbf{Document-level translation (d-chrF, $k=10$)} on \textsc{AfriDoc-MT} health domain with 3-shot prompting. \textit{Baseline}: Llama 3.1 8B SFT$_{10}$---fine-tuned on 4K documents from \textsc{AfriDoc-MT}  \cite{AFRIDOC-MT}.}
\label{tab:doc_level_mt}
\end{table}

\begin{table*}[t]
\centering
\footnotesize
\resizebox{\linewidth}{!}{%
\begin{tabular}{l|ccccccc|ccc}
\toprule
\textbf{Model} & \textbf{AfriMGSM} & \textbf{AfriMMLU} & \textbf{AfriXNLI} & \textbf{Belebele} & \textbf{Flores} & \textbf{Injongo} & \textbf{SIB-200} & \textbf{Overall} & \textbf{$\Delta$} & \textbf{$\Delta$ \%} \\
\midrule
\multicolumn{11}{c}{\texttt{Eval. on 20 most-resource African languages initially pre-trained}} \\
\midrule
Qwen 3 4B & 8.26 & 33.84 & 37.12 & 41.50 & 20.16 & 21.69 & 57.88 & 31.49 & - & - \\
\rowcolor[HTML]{E8F5E9} AfriqueQwen-4B & 33.09 & 43.04 & \textbf{44.88} & 63.62 & 59.82 & 65.34 & 74.77 & 54.94 & \textcolor{purple}{+23.4} & \textcolor{purple}{+74.4\%} \\
\midrule
Qwen 3.5 4B & 20.79 & 38.63 & 40.36 & 55.82 & 32.06 & 59.43 & 74.96 & 46.01 & - & - \\
\rowcolor[HTML]{E8F5E9} AfriqueQwen3.5-4B & 30.47 & 43.66 & 41.05 & 66.01 & 63.55 & 75.46 & 79.66 & 57.12 & \textcolor{purple}{+11.1} & \textcolor{purple}{+24.2\%} \\
\rowcolor[HTML]{E8F5E9} \; + ExtendedCM & \textbf{34.17} & \textbf{45.26} & 41.94 & 66.45 & 63.76 & \textbf{75.97} & \textbf{80.52} & \textbf{58.30} & \textcolor{purple}{+1.2}$^\dagger$ & \textcolor{purple}{+2.1\%}$^\dagger$ \\
\rowcolor[HTML]{E8F5E9}  \ \ \ \ \ \ \; + 30 Langs & 34.06 & 45.23 & 41.79 & \textbf{66.83} & \textbf{64.56} & 75.82 & 79.83 & \textbf{58.30} & \textcolor{purple}{+0.0}$^\ddagger$ & \textcolor{purple}{+0.0\%}$^\ddagger$ \\
\midrule
\multicolumn{11}{c}{\texttt{Eval. on 30 newly added languages, when covered by the benchmark}} \\
\midrule
Qwen 3.5 4B & 8.30 & 32.35 & 34.30 & 36.90 & 22.20 & 33.27 & 58.96 & 32.33 & - & - \\
\rowcolor[HTML]{E8F5E9} AfriqueQwen3.5-4B & 7.52 & 31.17 & 33.56 & 35.82 & 26.28 & 33.10 & 56.66 & 32.02 & \textcolor{purple}{-0.31} & \textcolor{purple}{-1.0\%} \\
\rowcolor[HTML]{E8F5E9} \; + ExtendedCM & 8.13 & 32.72 & 33.74 & 36.76 & 24.06 & 32.41 & 56.00 & 31.97 & \textcolor{purple}{-0.05}$^\dagger$ & \textcolor{purple}{-0.2\%}$^\dagger$ \\
\rowcolor[HTML]{E8F5E9}  \ \ \ \ \ \ \; + 30 Langs & \textbf{21.07} & \textbf{37.65} & \textbf{36.31} & \textbf{51.56} & \textbf{56.33} & \textbf{61.37} & \textbf{75.75} & \textbf{48.58} & \textcolor{purple}{+16.61}$^\ddagger$ & \textcolor{purple}{+52.0\%}$^\ddagger$ \\
\bottomrule
\end{tabular}
}
\vspace{-1.5mm}
\caption{\textbf{Qwen 3 family ablation: Effect of a more multilingual base (Qwen 3.5), 5$\times$ increase in Code/Math tokens (ExtendedCM), and expansion to 50 African languages.} $\Delta$ and $\Delta$ denote absolute and relative gains over the corresponding
base model. 
\textbf{Bold} marks the best among 4B Qwen Afrique variants within each evaluation block. $^\dagger$Relative to AfriqueQwen3.5-4B. $^\ddagger$Relative to AfriqueQwen3.5-4B-ExtendedCM.}
\label{tab:4b_qwen}
%\vspace{-1em}
\end{table*}

\subsection{Document-Level Translation}
\label{sec:doc-level-mt}

To evaluate whether our models with 16K tokens sequence length improve long-context translation, we benchmark on \textsc{AfriDoc-MT}~\cite{AFRIDOC-MT} (health domain), a document-level parallel corpus covering English and five African languages (Amharic, Hausa, Swahili, Yoruba, Zulu). We use pseudo-documents with $k=10$ sentences and report document-level chrF (d-chrF) scores with 3-shot prompting.
\autoref{tab:doc_level_mt} shows that all AfriqueLLMs excel at document-level translation despite never seeing \textsc{AfriDoc-MT} training data during CPT. We compare performance to Llama 3.1 SFT$_{10}$ baseline that was instruction fine-tuned on 4,060 health documents (812 per language pair).

For \textit{eng$\rightarrow$xx}, AfriqueGemma-12B achieves the best average (60.2), outperforming the task-specific SFT model (47.8) by +12.4 points. AfriqueQwen-14B (59.2) and AfriqueLlama-8B (58.5) also substantially exceed the SFT baseline, demonstrating that CPT provides robust long-context translation capabilities.

For \textit{xx$\rightarrow$eng}, AfriqueQwen-14B leads with 73.5, closely followed by AfriqueGemma-12B (73.3). Notably, both surpass the task-specific SFT$_{10}$ model (68.0), showing that CPT's general-purpose training can even exceed in-domain fine-tuning for certain translation directions.

\subsection{Qwen 3 Family: Impact of a More Multilingual Base Model}
\label{sec:4b-qwen}

To investigate the interplay between base-model multilingual coverage and data-mixture composition at smaller scale, the Qwen 3 and 3.5 series provide an ideal comparison, as they share the same architecture but differ in multilingual pre-training scope. Table~\ref{tab:4b_qwen} compares Qwen 3 4B (limited multilingual coverage i.e. only includes Afrikaans, Swahili, and Northern Arabic dialects) with Qwen 3.5 4B (expanded language support).\footnote{We hypothesize that the Qwen 3.5 series potentially includes more African languages than Qwen 3, given its support for 201 vs. 109 languages, and that it achieves better benchmark performance.}

Switching from Qwen 3 4B to the more multilingual Qwen 3.5 4B base substantially raises the starting point (31.49 $\rightarrow$ 46.01), which translates to a higher absolute post-CPT score (AfriqueQwen3.5-4B: 57.12 vs.~AfriqueQwen-4B: 54.94) but a smaller relative gain (+24.2\% vs.~+74.4\%). The stronger multilingual prior improves translation (Flores: 63.55 vs.~59.82) and language understanding tasks (AfriMMLU: 43.66 vs.~43.04), yet slightly weakens math-focused AfriMGSM (30.47 vs.~33.09). Increasing the Code and Math budget from 1B+1B tokens to 5B+5B tokens (\texttt{ExtendedCM}) largely recovers this gap on AfriMGSM (30.47 $\rightarrow$ 34.17) while also yielding consistent gains on AfriMMLU (43.66 $\rightarrow$ 45.26) and AfriXNLI (41.05 $\rightarrow$ 41.94), pushing the overall score to 58.26, comparable to AfriqueGemma-12B (58.82) at one-third the parameter count. 

This confirms that strong multilingual priors primarily benefit language-knowledge-intensive tasks such as translation---a pattern observed in both the Gemma 3 series and Qwen 3.5 4B. Meanwhile, the drop in math reasoning can be recovered by adding more domain-specific data. Notably, \textsc{ExtendedCM} improves most metrics while incurring only a negligible decrease on Flores, demonstrating that Code and Math data remain broadly beneficial across tasks, consistent with the findings from our data mixture ablation (Section~\ref{sec:data-mixture}).

% \paragraph{Scaling Language Coverage to 50 African Languages}
Expanding the CPT corpus from 20 to 50 African languages (\texttt{+30Langs}) broadens coverage without sacrificing the original set: on the 20 most-resourced languages, \texttt{+30Langs} matches \texttt{ExtendedCM} exactly (58.30 vs.~58.30), with slight redistribution of capacity (AfriMGSM 34.17 $\rightarrow$ 34.06) offset by translation gains (Flores: 63.76 $\rightarrow$ 64.56). On evaluation of the newly added languages in CPT, where benchmarks are available, both AfriqueQwen3.5-4B and its \texttt{ExtendedCM} variant remain at near base-model performance (32.02 and 31.97, respectively), indicating essentially no transfer. In contrast, explicitly including these languages during training increases the overall score to 48.58 (+16.61 points, +52.0\%), with the largest gains on translation- and knowledge-intensive tasks (Flores: 24.06 $\rightarrow$ 56.33; Injongo: 32.41 $\rightarrow$ 61.37). We report results only for the newly added languages covered by AfroBench-Lite (Ewe, Lingala, Luganda, Twi, and Wolof; Belebele covers only Lingala, Luganda, and Wolof), as detailed in Appendix~\ref{app:newly_added_langs}. These results reinforce that strong multilingual priors alone provide negligible transfer to unseen African languages, whereas a modest amount of in-language data unlocks substantial gains without degrading performance on previously mastered languages.

\section{Conclusion}
We introduce \textbf{AfriqueLLM}, a suite of LLMs adapted for 20 African languages via efficient CPT on 26B tokens. Our key findings are: \textit{(1)} data mixture matters most---combining monolingual text with code, math, and synthetic data (CMS) yields state-of-the-art results while preserving reasoning; \textit{(2)} base model strong capability trumps multilingual coverage---Qwen 3, despite minimal African language support, achieves the highest performance after CPT, with AfriqueQwen-14B (63.58) outperforming Gemma 3 27B (60.71) at less than half the size; and \textit{(3)} high-quality synthetic data provides a scalable bridge for low-resource languages, with AfriqueQwen-14B surpassing the 54B NLLB-MoE on translation. We have release our models on HuggingFace Model Hub to advance African language AI research.

% As future work, we plan to extend CPT to larger models in the Qwen 3.5 series and expand AfriqueLLM’s language coverage from its initial set to over 50 African languages.
%we plan to conduct a more comprehensive exploration and analysis on why Qwen 3 series models provides such a strong improvement after CPT, than similar architectures such as Gemma 3. 

% \section{}
% TODO: add endding for releasing the qwen3.5

\section{Limitations}
\label{sec:limitations}

\paragraph{Scope Constraints.}
Our study has several coverage limitations: (1) \textit{Language coverage}: We cover 20 African languages, leaving hundreds unsupported---languages with minimal digital presence remain challenging. (2) \textit{Model scale}: Resource constraints limited experiments to 14B parameters; larger models (30B+) may exhibit different adaptation dynamics and potential better performance, like ``Qwen3-30B-A3B-Base'' and ``Gemma 3 27b PT'' and potential better performance, like ``Qwen3-30B-A3B-Base'' and ``Gemma 3 27b PT''. (3) \textit{Training stage}: We focus on base model CPT without instruction tuning---the scarcity of high-quality instruction data for African languages remains a bottleneck for downstream deployment. (4) \textit{Hyperparameters}: Scaling to 12B+ prevents exhaustive search; we relied on heuristics from smaller model, which may not be optimal across architectures.

\paragraph{Training Stability and Efficiency.}
We observed intermittent gradient norm spikes during training, suggesting latent optimization instabilities. While these did not cause divergence, future work could explore matrix optimizers like Muon~\cite{Liu2025} for improved stability. Our framework achieves 31--34\% Model FLOPs Utilization (Appendix~\ref{app:efficiency}), competitive for general-purpose setups but leaving room for improvement via specialized frameworks like Megatron-LM~\cite{megatron-lm}.

\section{Acknowledgment}
This research was supported in part by the Natural Sciences and Engineering Research Council (NSERC) of Canada and in part by the AI2050 program at Schmidt Sciences. This work was partially supported by Azure sponsorship credits granted by Microsoft’s AI for Good Research Lab. We are grateful for the support from IVADO and the Canada First Research Excellence Fund.

%\newpage
\bibliography{afrobench,custom,related}

\appendix

\newpage
\newpage
\appendix

\section{Data Details}
\label{app:data_details}

\subsection{Language Selection and Statistics}
\label{app:lang_stats}

\begin{table*}[hbtp]
\centering
% \small

\begin{adjustbox}{max width=0.91\textwidth}
\begin{tabular}{@{}llrrrrcrrrr@{}}
\toprule
\textbf{Language} & \textbf{Code} & {\textbf{FineWeb2}} & {\textbf{Wura}} & {\textbf{Madlad400}} & {\textbf{Total Token}} & {\textbf{Rep.}} & {\textbf{Unimax Token}} & {\textbf{Synthetic}} & {\textbf{Other}} \\
\midrule
\multicolumn{10}{l}{\textbf{\textit{High-Resource (Non-African)} --- Capped at 1B tokens}} \\
English & eng\_Latn & $>$1000000000 & 865600280 & {--} & 1000000000 & 1$\times$ & 1070793848 & 16,011,265 & \\
French & fra\_Latn & $>$1000000000 & 815336425 & {--} & 1000000000 & 1$\times$ & 1070793848 & & \\
Portuguese & por\_Latn & $>$1000000000 & 531069643 & {--} & 1000000000 & 1$\times$ & 1070793849 & & \\
Arabic & arb\_Arab & $>$1000000000 & {--} & {--} & 1000000000 & 1$\times$ & 1070793848 & & \\
\multicolumn{10}{l}{\textbf{\textit{African Languages} --- Included in Training}} \\
Afrikaans & afr\_Latn & 2461214686 & 1357859486 & 1483495285 & 5302569457 & 1$\times$ & 1070793849 & 12,113,273 & \\
Swahili & swh\_Latn & 1051220388 & 1087449729 & 777825674 & 2916495791 & 1$\times$ & 1070793849 & 12,503,168 & \\
Moroccan Arabic & ary\_Arab & 3289564375 & {--} & {--} & 3289564375 & 1$\times$ & 1070793849 & & \\
Somali & som\_Latn & 732191814 & 702650753 & 346518006 & 1781360573 & 1$\times$ & 1070793849 & 13,572,904 & \\
Amharic & amh\_Ethi & 403784914 & 276855513 & 308253510 & 988893937 & 2$\times$ & 1070793848 & 23,943,363 & \\
Egyptian Arabic & arz\_Arab & 821465539 & 131515140 & {--} & 952980679 & 2$\times$ & 1070793848 & & \\
Hausa & hau\_Latn & {--} & 288353911 & 211672798 & 500026709 & 3$\times$ & 1070793848 & 12,596,581 & \\
Kinyarwanda & kin\_Latn & 136010710 & 69028912 & 275720285 & 480759907 & 3$\times$ & 1070793848 & 12,707,048 & \\
Zulu & zul\_Latn & 159037587 & 97653578 & 92982744 & 349673909 & 4$\times$ & 1070793848 & 12,366,125 & \\
Igbo & ibo\_Latn & 140734354 & 68796722 & 108189914 & 317720990 & 4$\times$ & 1070793848 & 12,671,171 & \\
Plateau Malagasy & plt\_Latn & 310443854 & {--} & {--} & 310443854 & 4$\times$ & 1070793848 & 14,002,182 & \\
Xhosa & xho\_Latn & 119027393 & 41737419 & 107219367 & 267984179 & 4$\times$ & 1070793848 & 14,846,741 & \\
Shona & sna\_Latn & 95516967 & 76561301 & 90581980 & 262660248 & 5$\times$ & 1050640992 & 10,971,211 & \\
Yoruba & yor\_Latn & 90126934 & 68250903 & 99303113 & 257680950 & 5$\times$ & 1030723800 & 17,152,436 & \\
Nyanja & nya\_Latn & 137607319 & 92652643 & {--} & 230259962 & 5$\times$ & 921039848 & 11,481,563 & \\
Southern Sotho & sot\_Latn & 122964390 & {--} & 80276553 & 203240943 & 5$\times$ & 812963772 & 13,573,191 & \\
Tigrinya & tir\_Ethi & 100865939 & 8661533 & 32703052 & 142230524 & 5$\times$ & 568922096 & 75,525,088 & \\
Tunisian Arabic & aeb\_Arab & 136652951 & {--} & {--} & 136652951 & 5$\times$ & 546611804 & & \\
West Central Oromo & gaz\_Latn & 42916258 & 17619689 & 32493752 & 93029699 & 5$\times$ & 372118796 & 21,619,016 & \\
Tswana & tsn\_Latn & 9244373 & 72533425\footnote{From HuggingFace dataset OxxoCodes/Marothodi, not WURA.} & 10215596 & 91993394 & 5$\times$ & 367973576 & 16,313,360 & \\
\midrule
\multicolumn{10}{l}{\textbf{\textit{Additional Training Data}}} \\
FineMath (Math, M) & {--} & {--} & {--} & {--} & {--} & {--} & {--} & {--} & 1,067,549,046 \\
CornStack-Python (Code, C) & {--} & {--} & {--} & {--} & {--} & {--} & {--} & {--} & 967,399,767 \\
MT-NLLB (Parallel, P) & {--} & {--} & {--} & {--} & {--} & {--} & {--} & {--} & 456,102,720 \\
\midrule
\multicolumn{5}{l}{\textbf{Subtotal of Tokens}} & \textbf{22.88B} & & \textbf{22.80B} & \textbf{0.32B} & \textbf{2.49B} \\
\midrule
\multicolumn{10}{l}{\textbf{\textit{Newly Added Languages} --- Trained only in the 50-Language Version}} \\
Rundi & run\_Latn & 56775951 & {--} & 492969 & 57268920 & 5$\times$ & 286344600 & & \\
Ganda & lug\_Latn & 24162781 & {--} & 18022976 & 42185757 & 5$\times$ & 210928785 & & \\
Tsonga & tso\_Latn & 10782436 & {--} & 14451048 & 25233484 & 5$\times$ & 126167420 & & \\
Lingala & lin\_Latn & 16358800 & {--} & 7530450 & 23889250 & 5$\times$ & 119446250 & & \\
Ewe & ewe\_Latn & 3014541 & {--} & 15388319 & 18402860 & 5$\times$ & 92014300 & & \\
Wolof & wol\_Latn & 16527037 & {--} & 1839642 & 18366679 & 5$\times$ & 91833395 & & \\
Sango & sag\_Latn & 7104619 & {--} & 5590802 & 12695421 & 5$\times$ & 63477105 & & \\
Akan & aka\_Latn & {--} & {--} & 10824690 & 10824690 & 5$\times$ & 54123450 & & \\
Twi & twi\_Latn & 10648719 & {--} & {--} & 10648719 & 5$\times$ & 53243595 & & \\
Kabiye & kbp\_Latn & 1040478 & {--} & 8959130 & 9999608 & 5$\times$ & 49998040 & & \\
Bambara & bam\_Latn & 7335041 & {--} & 1426843 & 8761884 & 5$\times$ & 43809420 & & \\
Northern Sotho & nso\_Latn & 8630368 & {--} & {--} & 8630368 & 5$\times$ & 43151840 & & \\
Fon & fon\_Latn & 2281350 & {--} & 4439623 & 6720973 & 5$\times$ & 33604865 & & \\
Swati & ssw\_Latn & 2660736 & {--} & 2016953 & 4677689 & 5$\times$ & 23388445 & & \\
Tamazight & tzm\_Tfng & 4044801 & {--} & 260465 & 4305266 & 5$\times$ & 21526330 & & \\
Kabyle & kab\_Latn & 3860016 & {--} & {--} & 3860016 & 5$\times$ & 19300080 & & \\
Kabuverdianu & kea\_Latn & 3782732 & {--} & {--} & 3782732 & 5$\times$ & 18913660 & & \\
N'ko & nqo\_Nkoo & 3717948 & {--} & {--} & 3717948 & 5$\times$ & 18589740 & & \\
Mossi & Mos\_Latn & 3319912 & {--} & {--} & 3319912 & 5$\times$ & 16599560 & & \\
Kimbundu & kmb\_Latn & 1506689 & {--} & 1759056 & 3265745 & 5$\times$ & 16328725 & & \\
Kanuri (Arabic) & knc\_Arab & 3105431 & {--} & {--} & 3105431 & 5$\times$ & 15527155 & & \\
Kanuri (Latin) & knc\_Latn & 256317 & {--} & {--} & 256317 & 5$\times$ & 1281585 & & \\
Dyula & dyu\_Latn & 2018490 & {--} & 960718 & 2979208 & 5$\times$ & 14896040 & & \\
Tamasheq (Latin) & taq\_Latn & 2640160 & {--} & {--} & 2640160 & 5$\times$ & 13200800 & & \\
Southwestern Dinka & dik\_Latn & 1144214 & {--} & 1420754 & 2564968 & 5$\times$ & 12824840 & & \\
Luo & luo\_Latn & 2010521 & {--} & {--} & 2010521 & 5$\times$ & 10052605 & & \\
Nigerian Fulfulde & fuv\_Latn & 1894553 & {--} & 95651 & 1990204 & 5$\times$ & 9951020 & & \\
Bemba & bem\_Latn & 1482559 & {--} & {--} & 1482559 & 5$\times$ & 7412795 & & \\
Kikuyu & kik\_Latn & 1411871 & {--} & {--} & 1411871 & 5$\times$ & 7059355 & & \\
Kamba & kam\_Latn & 1018287 & {--} & {--} & 1018287 & 5$\times$ & 5091435 & & \\
Kikongo & kon\_Latn & {--} & {--} & 971858 & 971858 & 5$\times$ & 4859290 & & \\
Luba-Kasai & lua\_Latn & 908010 & {--} & {--} & 908010 & 5$\times$ & 4540050 & & \\
\textbf{Newly Added Subtotal} &  &  &  &  & \textbf{301,897,315} &  & \textbf{1,509,486,575} &  & \\
\midrule
\multicolumn{10}{l}{\textbf{\textit{Excluded Languages} --- Not trained in any version}} \\
Umbundu & umb\_Latn & 540735 & {--} & {--} & 540735 & 0 & 0 & & \\
Tamasheq (Tifinagh) & taq\_Tfng & 401256 & {--} & {--} & 401256 & 0 & 0 & & \\
Tumbuka & tum\_Latn & 228626 & {--} & {--} & 228626 & 0 & 0 & & \\
Nuer & nus\_Latn & 224103 & {--} & {--} & 224103 & 0 & 0 & & \\
Chokwe & cjk\_Latn & 33366 & {--} & {--} & 33366 & 0 & 0 & & \\
\textbf{Excluded Subtotal} &  &  &  &  & \textbf{1,428,086} &  & \textbf{0} &  & \\
\bottomrule
\end{tabular}
\end{adjustbox}
\caption{Complete dataset collection and language selection for training. This table presents all 60+ African and high-resource languages collected from FineWeb2, Wura, and Madlad400 sources, along with the selection criteria applied. Languages with $\geq$90M tokens are included in the base training set (24 languages, 22.8B tokens). The remaining lower-resource languages are incorporated \emph{only} in the \textsc{+50Langs} version, where their UniMax-upsampled counts appear under ``Unimax Token''; Umbundu, Tamasheq (Tifinagh), Tumbuka, Nuer, and Chokwe are not trained in any version. The ``Rep.'' column indicates the upsampling factor applied via UniMax to balance low-resource languages.}
\label{tab:dataset-detail}
\end{table*}

Table~\ref{tab:dataset-detail} provides a comprehensive overview of the language selection process and the final token counts for each language across our primary data sources: FineWeb2, WURA, and MADLAD-400. We applied a selection threshold of 90M tokens to ensure sufficient data for meaningful linguistic adaptation.

\subsection{Synthetic Data and Translation Prompts}
\label{app:synthetic_details}

Table~\ref{tab:synthetic_dist} details the distribution of synthetic data across 11 domains. The translation process was guided by the prompts shown in Section~\ref{app:translation_prompts}.

\begin{table}[h]
\centering
\small

\begin{tabular}{lr}
\toprule
\textbf{Domain} & \textbf{Tokens} \\
\midrule
Math & 32,284,225 \\
Science \& Tech. & 37,461,084 \\
Politics & 35,256,194 \\
Health & 31,213,028 \\
Travel & 29,751,012 \\
History & 28,386,610 \\
Food \& Dining & 27,556,953 \\
Education \& Jobs & 27,469,250 \\
Software Dev. & 26,446,379 \\
Entertainment & 25,148,472 \\
Industrial & 22,996,479 \\
\midrule
\textbf{Total} & \textbf{323,969,686} \\
\bottomrule
\end{tabular}
\caption{Synthetic Data domain distribution. Math data is sourced from \cite{OpenMathReasoning}, while other domains are from \cite{OrganizeWeb}.}
\label{tab:synthetic_dist}
\end{table}

\label{app:translation_prompts}
\begin{codeblock}{General Translation Prompt}
\begin{lstlisting}[basicstyle=\ttfamily\footnotesize, breaklines=true]
You are a professional translator. Translate the user text from {{source_lang}} into {{target_lang}}. 
Preserve meaning, tone, formatting, inline markup, numerals, and named entities exactly. 
For long texts, ensure the translation is fluent, coherent and complete. Make sure to translate all parts of the text. Return only the translation without additional commentary.
\end{lstlisting}
\end{codeblock}

\begin{codeblock}{Mathematical Reasoning Translation Prompt}
\begin{lstlisting}[basicstyle=\ttfamily\footnotesize, breaklines=true]
You are a {{source_lang}}-to-{{target_lang}} translator for mathematical content. Translate the provided math problem, reasoning, and answer while preserving:
- All numbers, formulas, and formatting
- Mathematical notation and markup
- Named entities and tone

Input structure:
<problem>[Original Problem]</problem>
<think>[Original Reasoning]</think>
[Final Answer]<eos>

Output structure:
<problem>[Translated problem]</problem>
<think>[Translated reasoning]</think>
[Translated Final Answer]<eos>

Ensure translations are fluent, coherent, and complete. Return only the translation without additional commentary.
\end{lstlisting}
\end{codeblock}

\section{Training Details}
\label{app:training_details}

\subsection{Hyperparameter Search}
\label{app:hp_search}

We conducted an extensive ablation study to identify the optimal hyperparameters for continued pre-training on African languages.

\paragraph{Learning Rate}
We performed a learning rate sweep on the Gemma 3 4B PT model with rates ranging from $1\text{e-}6$ to $2\text{e-}4$. Table~\ref{tab:lr_search} identifies $5\text{e-}5$ as the optimal rate based on average scores across low-resource languages.

\begin{table}[h]
\centering
\small

\begin{adjustbox}{max width=\columnwidth}
\begin{tabular}{l|cccc}
\toprule
\textbf{LR} & \textbf{AfriMGSM} & \textbf{AfriXNLI} & \textbf{AfriMMLU} & \textbf{Flores} \\
\midrule
2e-4 & 5.1 & 39.0 & 27.5 & - \\
1e-4 & 8.1 & 38.6 & 31.0 & - \\
5e-5 & \textbf{9.4} & \textbf{40.6} & 34.7 & \textbf{61.1} \\
2e-5 & 8.4 & 40.5 & 36.0 & 59.4 \\
1e-5 & 8.0 & 39.9 & \textbf{37.2} & 57.5 \\
5e-6 & 8.6 & 39.4 & 36.7 & 54.6 \\
2e-6 & 9.0 & 37.8 & 36.8 & 50.3 \\
1e-6 & 8.7 & 38.7 & 36.4 & 46.4 \\
\bottomrule
\end{tabular}
\end{adjustbox}
\caption{Ablation on learning rate. Results are reported for the Gemma 3 4B pretrained model with a fixed 16k context length. We report average scores for AfriMGSM (8-shot CoT), AfriXNLI (Direct), AfriMMLU (Direct), and Translation (SSA-COMET) excluding English and French.}
\label{tab:lr_search}
\end{table}

\paragraph{Context Size}
Using the optimal learning rate ($5\text{e-}5$), we evaluated context lengths of 4k, 16k, and 32k. Table~\ref{tab:context_size_search} shows that a 16k context window yields the best performance on AfriMGSM.

\begin{table}[h]
\centering
\small

\begin{adjustbox}{max width=\columnwidth}
\begin{tabular}{l|c}
\toprule
\textbf{Context Length} & \textbf{AfriMGSM} \\
\midrule
4k & 7.5 \\
16k & \textbf{9.4} \\
32k & 7.8 \\
\bottomrule
\end{tabular}
\end{adjustbox}
\caption{Ablation on context length. Results are reported for the Gemma 3 4B pretrained model with a fixed 5e-5 learning rate. We report average scores for AfriMGSM (8-shot CoT), excluding English and French.}
\label{tab:context_size_search}
\end{table}

% TODO: forget to use the best cosine scheduler
\paragraph{Cosine Scheduler} 
We explored the impact of the minimum learning rate (min lr) and warmup steps. Table~\ref{tab:scheduler_search} presents the results using the Gemma 3 4B pretrained model with a fixed context size of 16k.

\begin{table}[h]
\centering
\small
\begin{adjustbox}{max width=\columnwidth}
\begin{tabular}{lc|cccc}
\toprule
\textbf{Min lr} & \textbf{Warmup} &\textbf{AfriMGSM} & \textbf{AfriXNLI} & \textbf{AfriMMLU} & \textbf{Flores}\\
\midrule
0.01 & 0 & 9.4 & \textbf{38.8} & \textbf{34.1} & 60.6 \\
0.01 & 0.001 & \textbf{10.2} & 38.2 & \textbf{34.1} & 60.5 \\
0.1 & 0 & 7.7 & 38.5 & 34.0 & 60.9 \\
0.1 & 0.001 & 9.4 & 38.5 & 33.4 & \textbf{61.1} \\
\bottomrule
\end{tabular}
\end{adjustbox}
\caption{Ablation on cosine scheduler hyperparameters. Results are reported for the Gemma 3 4B pretrained model with a 16k context window. We report average scores for AfriMGSM (8-shot CoT), AfriXNLI (Direct), AfriMMLU (Direct), and Translation (SSA-COMET) excluding English and French.}
\label{tab:scheduler_search}
\end{table}

\subsection{Training Configuration}
\label{app:training_config}

The following YAML configuration was used for the continued pre-training of the AfriqueLLM models using the LLaMA-Factory framework.

\begin{codeblock}{LLaMA-Factory CPT config (YAML)}
\small
\begin{lstlisting}[language=yaml]
### model
model_name_or_path: google/gemma-3-12b-pt

### method
stage: pt

### data
template: empty
packing: true
cutoff_len: 16384 # 16k
overwrite_cache: false
preprocessing_num_workers: 32
dataloader_num_workers: 32

### finetuning
do_train: true
finetuning_type: full
deepspeed: ds_z1_config.json
freeze_vision_tower: true
freeze_multi_modal_projector: true
freeze_language_model: false

### output
logging_steps: 10
save_steps: 1000
plot_loss: true
overwrite_output_dir: true
save_only_model: false
report_to: wandb
data_shared_file_system: true

### train
per_device_train_batch_size: 4
gradient_accumulation_steps: 8
learning_rate: 5.0e-5
num_train_epochs: 1.0
lr_scheduler_type: cosine_with_min_lr
lr_scheduler_kwargs: 
  min_lr_rate: 0.01
warmup_ratio: 0.001
bf16: true
ddp_timeout: 180000000
resume_from_checkpoint: null
weight_decay: 0.1
adam_beta1: 0.9
adam_beta2: 0.95

enable_liger_kernel: true
flash_attn: fa3
\end{lstlisting}
\end{codeblock}

\newpage
\subsection{Training Efficiency Analysis}
\label{app:efficiency}

Table~\ref{tab:efficiency} summarizes the computational metrics for our continued pre-training process.

\begin{table}[h]
\centering
\small

\begin{adjustbox}{max width=\linewidth}
\begin{tabular}{lccccccccc}
\toprule
\textbf{Model} & \textbf{Nodes} & \textbf{GPUs} & \textbf{Steps} & \textbf{FLOPs} & \textbf{Time (h)} & \textbf{TFLOPS} & \textbf{MFU (\%)} & \textbf{Loss} \\
\midrule
AfriqueGemma 4B & 4 & 16 & 6,008 & 0.55 ZFLOPs & 9.12 & 16,690 & 26.37 & 1.5174 \\
AfriqueGemma 12B & 16 & 64 & 6,000 & 1.69 ZFLOPs & 23.70 & 19,776 & 31.24 & 1.2942 \\
AfriqueQwen 8B & 16 & 64 & 6,872 & 1.31 ZFLOPs & 18.30 & 19,868 & 31.39 & 1.3375 \\
AfriqueQwen 14B & 16 & 64 & 6,872 & 2.42 ZFLOPs & 31.10 & 21,622 & 34.16 & 1.1865 \\
AfriqueLlama 8B & 16 & 64 & 7,406 & 1.40 ZFLOPs & 18.06 & 21,516 & 33.99 & 1.1355 \\
\bottomrule
\end{tabular}
\end{adjustbox}
\caption{Training efficiency metrics for Continued Pre-Training (CPT) on H100 GPUs. FLOPs indicates total floating-point operations. Loss refers to the final step training loss.}
\label{tab:efficiency}
\end{table}

\section{Evaluation Details}
\label{app:evaluation_details}

\subsection{Benchmark Language Coverage}
\label{app:bench_coverage}

Table~\ref{tab:benchmarks-langs} lists the languages covered in each task of the AfroBench-Lite suite.

\begin{table}[h]
\centering
\small
\begin{tabular}{p{0.07\textwidth}p{0.39\textwidth}}
\toprule
\textbf{Task} & \textbf{Languages (Total Counts)} \\
\midrule
\textsc{afrimgsm} & Amharic$^\dagger$, English$^*$, Ewe, French$^*$, Hausa$^\dagger$, Igbo$^\dagger$, Kinyarwanda$^\dagger$, Lingala, Luganda, Oromo$^\dagger$, Shona$^\dagger$, Sotho$^\dagger$, Swahili$^\dagger$, Twi, Vai, Wolof, Xhosa$^\dagger$, Yoruba$^\dagger$, Zulu$^\dagger$ (19)\\
\textsc{afrimmlu} & Amharic$^\dagger$, English$^*$, Ewe, French$^*$, Hausa$^\dagger$, Igbo$^\dagger$, Kinyarwanda$^\dagger$, Lingala, Luganda, Oromo$^\dagger$, Shona$^\dagger$, Sotho$^\dagger$, Swahili$^\dagger$, Twi, Wolof, Xhosa$^\dagger$, Yoruba$^\dagger$, Zulu$^\dagger$ (18)\\
\textsc{afrixnli} & Amharic$^\dagger$, English$^*$, Ewe, French$^*$, Hausa$^\dagger$, Igbo$^\dagger$, Kinyarwanda$^\dagger$, Lingala, Luganda, Oromo$^\dagger$, Shona$^\dagger$, Sotho$^\dagger$, Swahili$^\dagger$, Twi, Wolof, Xhosa$^\dagger$, Yoruba$^\dagger$, Zulu$^\dagger$ (18)\\
\textsc{belebele} & Afrikaans$^\dagger$, Amharic$^\dagger$, Egyptian Arabic$^\dagger$, English$^*$, French$^*$, Hausa$^\dagger$, Igbo$^\dagger$, Kinyarwanda$^\dagger$, Lingala, Luganda, Moroccan Arabic$^\dagger$, Nyanja$^\dagger$, Oromo$^\dagger$, Plateau Malagasy$^\dagger$, Portuguese$^*$, Shona$^\dagger$, Somali$^\dagger$, Sotho$^\dagger$, Swahili$^\dagger$, Tigrinya$^\dagger$, Tswana$^\dagger$, Wolof, Xhosa$^\dagger$, Yoruba$^\dagger$, Zulu$^\dagger$ (25) \\
\textsc{flores} & Afrikaans$^\dagger$, Amharic$^\dagger$, Egyptian Arabic$^\dagger$, Ewe, Hausa$^\dagger$, Igbo$^\dagger$, Kinyarwanda$^\dagger$, Lingala, Luganda, Moroccan Arabic$^\dagger$, Nyanja$^\dagger$, Oromo$^\dagger$, Shona$^\dagger$, Somali$^\dagger$, Sotho$^\dagger$, Swahili$^\dagger$, Tigrinya$^\dagger$, Tswana$^\dagger$, Tunisian Arabic$^\dagger$, Twi, Wolof, Xhosa$^\dagger$, Yoruba$^\dagger$, Zulu$^\dagger$ (24) \\
\textsc{injongo} & Amharic$^\dagger$, English$^*$, Ewe, Hausa$^\dagger$, Igbo$^\dagger$, Kinyarwanda$^\dagger$, Lingala, Luganda, Oromo$^\dagger$, Shona$^\dagger$, Sotho$^\dagger$, Swahili$^\dagger$, Twi, Wolof, Xhosa$^\dagger$, Yoruba$^\dagger$, Zulu$^\dagger$ (17) \\
\textsc{sib-200} & Afrikaans$^\dagger$, Amharic$^\dagger$, Egyptian Arabic$^\dagger$, English$^*$, Ewe, Hausa$^\dagger$, Igbo$^\dagger$, Kinyarwanda$^\dagger$, Lingala, Luganda, Moroccan Arabic$^\dagger$, Nyanja$^\dagger$, Oromo$^\dagger$, Plateau Malagasy$^\dagger$, Portuguese$^*$, Shona$^\dagger$, Somali$^\dagger$, Sotho$^\dagger$, Swahili$^\dagger$, Tigrinya$^\dagger$, Tunisian Arabic$^\dagger$, Twi, Wolof, Xhosa$^\dagger$, Yoruba$^\dagger$, Zulu$^\dagger$ (26)\\
\bottomrule
\end{tabular}
\caption{Languages included in each benchmark task. \\ $^*$: High-resource pretrained (4) \\ $^\dagger$: Pretrained African (20)}
\label{tab:benchmarks-langs}
\end{table}

\subsection{Newly Added Languages in the 50-Language Version}
\label{app:newly_added_langs}

In the \textsc{+50Langs} configuration, we expand the CPT corpus from 20 to 50 African languages by including the previously held-out languages whose \emph{Final Used Tokens} are non-zero in Table~\ref{tab:dataset-detail} (i.e., languages retained after UniMax upsampling); languages with zero final tokens (e.g., Umbundu, Tumbuka, Nuer, Chokwe) remain excluded. Of these newly added languages, only a subset is covered by the AfroBench-Lite evaluation suite. Table~\ref{tab:newly-added-bench} lists, per benchmark, the newly added languages that are actually evaluated; these are the languages reported in the ``30 newly added languages'' block of Table~\ref{tab:4b_qwen}.

\begin{table}[h]
\centering
\small
\begin{tabular}{p{0.30\columnwidth}p{0.58\columnwidth}}
\toprule
\textbf{Benchmark} & \textbf{Newly Added Languages Evaluated} \\
\midrule
\textsc{afrimgsm} & Ewe, Lingala, Luganda, Twi, Wolof \\
\textsc{afrimmlu} & Ewe, Lingala, Luganda, Twi, Wolof \\
\textsc{afrixnli} & Ewe, Lingala, Luganda, Twi, Wolof \\
\textsc{belebele} & Lingala, Luganda, Wolof \\
\textsc{flores} (eng$\to$xx) & Ewe, Lingala, Luganda, Twi, Wolof \\
\textsc{flores} (xx$\to$eng) & Ewe, Lingala, Luganda, Twi, Wolof \\
\textsc{injongo} & Ewe, Lingala, Luganda, Twi, Wolof \\
\textsc{sib-200} & Ewe, Lingala, Luganda, Twi, Wolof \\
\bottomrule
\end{tabular}
\caption{Newly added languages (beyond the original 20) that are covered by each benchmark in the \textsc{+50Langs} evaluation. Only these languages contribute to the ``30 newly added languages'' evaluation block in Table~\ref{tab:4b_qwen}.}
\label{tab:newly-added-bench}
\end{table}

\newpage
\section{Detailed Experimental Results}
\label{app:detailed_results}

\begin{table*}[h]
\centering
\begin{adjustbox}{width=\textwidth}
\begin{tabular}{lcccccccccccccccccccc}
\toprule
model & amh & eng & ewe & fra & hau & ibo & kin & lin & lug & orm & sna & sot & swa & twi & vai & wol & xho & yor & zul & avg \\
\midrule
Llama3.1-8B & 2.72 & 53.52 & 3.44 & 37.12 & 13.12 & 7.76 & 6.64 & 4.40 & 7.28 & 4.80 & 6.80 & 6.64 & 23.84 & 5.44 & 1.84 & 5.04 & 4.00 & 6.64 & 6.56 & 10.93 \\
Lugha-Llama-8B-wura & 4.72 & 40.88 & 2.00 & 20.32 & 12.16 & 10.24 & 9.76 & 2.72 & 4.56 & 8.16 & 10.08 & 7.68 & 19.28 & 3.28 & 2.88 & 1.92 & 6.40 & 7.44 & 8.16 & 9.61 \\
AfriqueLlama-8B & 7.84 & 55.52 & 3.12 & 36.88 & 20.96 & 15.04 & 18.96 & 6.40 & 11.52 & 17.52 & 20.72 & 18.80 & 24.48 & 4.08 & 1.68 & 3.92 & 13.04 & 19.52 & 15.76 & 16.62 \\
\midrule
Gemma3-4B & 10.64 & 42.48 & 3.28 & 28.72 & 12.88 & 5.28 & 7.76 & 2.88 & 6.56 & 2.64 & 12.16 & 10.08 & 27.12 & 1.84 & 0.08 & 2.40 & 7.68 & 5.44 & 10.96 & 10.57 \\
AfriqueGemma-4B & 17.52 & 37.84 & 3.60 & 21.52 & 17.04 & 13.36 & 12.96 & 4.16 & 10.16 & 9.68 & 16.96 & 15.84 & 21.60 & 2.00 & 0.88 & 2.24 & 10.72 & 11.68 & 16.08 & 12.94 \\
\midrule
Gemma3-12B & 38.64 & 72.40 & 6.08 & 50.16 & 26.00 & 22.08 & 22.08 & 13.60 & 19.84 & 14.32 & 29.52 & 20.00 & 46.40 & 5.92 & 1.52 & 4.88 & 17.60 & 13.28 & 27.36 & 23.77 \\
AfriqueGemma-12B & 36.00 & 68.08 & 4.48 & 57.20 & 34.88 & 30.72 & 24.72 & 7.76 & 19.84 & 28.48 & 31.52 & 33.44 & 47.36 & 6.40 & 0.80 & 2.64 & 26.32 & 26.48 & 33.60 & 27.41 \\
\midrule
Qwen3-4B & 8.80 & 84.40 & 6.08 & 69.52 & 6.64 & 2.16 & 6.80 & 7.12 & 7.12 & 9.04 & 7.68 & 8.56 & 23.84 & 5.68 & 3.28 & 6.24 & 5.12 & 5.52 & 6.72 & 14.75 \\
AfriqueQwen-4B & 30.88 & 79.12 & 7.12 & 66.32 & 38.32 & 26.96 & 32.72 & 8.88 & 16.72 & 30.08 & 32.88 & 35.52 & 46.72 & 5.84 & 3.04 & 4.72 & 25.60 & 34.88 & 29.44 & 29.25 \\
\midrule
Qwen3.5-4B & 32.00 & 82.80 & 8.24 & 69.28 & 22.40 & 9.44 & 19.12 & 9.20 & 11.04 & 10.48 & 16.24 & 22.64 & 41.04 & 4.64 & 2.96 & 8.40 & 17.04 & 18.64 & 19.60 & 22.38 \\
AfriqueQwen3.5-4B & 30.16 & 75.36 & 3.76 & 55.44 & 37.20 & 27.04 & 32.00 & 8.64 & 16.96 & 29.68 & 30.80 & 27.84 & 43.68 & 4.40 & 1.76 & 3.84 & 22.00 & 27.84 & 26.96 & 26.60 \\
AfriqueQwen3.5-4B-ExtendedCM & 35.12 & 78.64 & 6.16 & 61.60 & 42.32 & 27.12 & 34.24 & 8.08 & 20.32 & 31.92 & 34.40 & 32.96 & 51.36 & 3.60 & 2.56 & 2.48 & 26.80 & 30.40 & 29.28 & 29.44 \\
AfriqueQwen3.5-4B-50Langs & 35.60 & 78.24 & 21.28 & 61.84 & 42.16 & 25.04 & 36.08 & 26.88 & 27.84 & 31.68 & 33.92 & 32.00 & 48.88 & 13.60 & 2.16 & 15.76 & 25.36 & 31.20 & 32.72 & 32.75 \\
\midrule
Qwen3-8B & 10.80 & 85.76 & 5.92 & 74.08 & 7.84 & 2.64 & 8.88 & 7.92 & 8.00 & 12.16 & 8.48 & 10.88 & 39.04 & 5.76 & 1.12 & 5.20 & 8.80 & 6.48 & 7.44 & 16.69 \\
AfriqueQwen-8B & 40.48 & 85.20 & 6.40 & 67.92 & 48.88 & 32.08 & 42.56 & 9.76 & 22.16 & 37.76 & 40.00 & 36.64 & 57.28 & 5.68 & 3.04 & 4.88 & 30.24 & 35.44 & 35.12 & 33.76 \\
\midrule
Qwen3-14B & 12.88 & 88.00 & 8.80 & 76.56 & 13.68 & 3.60 & 15.68 & 12.08 & 12.96 & 19.04 & 11.52 & 16.16 & 50.40 & 6.64 & 1.92 & 5.84 & 13.92 & 13.84 & 11.92 & 20.81 \\
\rowcolor[HTML]{E8F5E9}
AfriqueQwen-14B & 35.28 & 82.24 & 7.68 & 72.24 & 52.32 & 41.44 & 46.96 & 12.24 & 27.36 & 47.12 & 46.80 & 45.20 & 67.04 & 5.68 & 2.88 & 5.52 & 31.84 & 42.32 & 38.80 & 37.42 \\
\bottomrule
\end{tabular}
\end{adjustbox}
\caption{8-shot performance on the AfriMGSM benchmark for multilingual grade school math. (Math)}
\label{tab:afrimgsm}
\end{table*}

\begin{table*}[h]
\centering

\begin{adjustbox}{width=\textwidth}
\begin{tabular}{lccccccccccccccccccc}
\toprule
model & amh & eng & ewe & fra & hau & ibo & kin & lin & lug & orm & sna & sot & swa & twi & wol & xho & yor & zul & avg \\
\midrule
Llama3.1-8B & 34.16 & 65.56 & 27.48 & 50.64 & 33.84 & 31.72 & 32.80 & 34.84 & 30.84 & 32.24 & 29.84 & 29.24 & 39.08 & 28.48 & 30.20 & 27.76 & 32.08 & 32.20 & 34.61 \\
Lugha-Llama-8B-wura & 38.52 & 65.08 & 27.80 & 51.80 & 37.76 & 37.96 & 33.32 & 33.00 & 30.36 & 34.60 & 38.72 & 38.48 & 41.84 & 27.44 & 29.52 & 32.68 & 34.52 & 38.60 & 37.33 \\
AfriqueLlama-8B & 38.28 & 58.04 & 29.28 & 46.48 & 36.48 & 37.36 & 32.92 & 31.68 & 28.32 & 36.48 & 35.84 & 37.52 & 42.24 & 28.12 & 26.68 & 34.72 & 35.56 & 34.92 & 36.16 \\
\midrule
Gemma3-4B & 34.40 & 58.00 & 28.24 & 49.68 & 34.08 & 35.36 & 34.04 & 29.92 & 27.48 & 28.00 & 33.88 & 34.04 & 41.48 & 28.92 & 26.04 & 33.84 & 29.00 & 34.72 & 34.51 \\
AfriqueGemma-4B & 38.80 & 54.48 & 29.16 & 47.56 & 37.92 & 37.04 & 34.72 & 28.04 & 27.88 & 34.16 & 36.56 & 36.24 & 40.40 & 26.64 & 25.08 & 37.80 & 36.24 & 34.12 & 35.71 \\
\midrule
Gemma3-12B & 52.84 & 78.08 & 30.72 & 70.40 & 50.96 & 47.96 & 45.80 & 42.40 & 38.88 & 39.64 & 50.00 & 49.12 & 61.56 & 33.76 & 29.80 & 46.72 & 43.56 & 48.20 & 47.80 \\
AfriqueGemma-12B & 51.88 & 70.64 & 25.84 & 62.08 & 49.20 & 47.64 & 46.72 & 37.76 & 37.36 & 47.28 & 50.60 & 51.56 & 54.48 & 33.40 & 26.36 & 49.60 & 47.72 & 47.52 & 46.54 \\
\midrule
Qwen3-4B & 34.68 & 75.24 & 32.56 & 65.96 & 33.80 & 34.64 & 31.04 & 38.52 & 31.28 & 36.28 & 32.52 & 34.20 & 36.56 & 33.08 & 32.84 & 32.16 & 34.32 & 32.00 & 37.87 \\
AfriqueQwen-4B & 49.88 & 71.40 & 30.16 & 61.28 & 44.52 & 41.24 & 39.88 & 34.92 & 31.16 & 43.44 & 42.28 & 42.24 & 49.32 & 31.12 & 30.52 & 38.80 & 40.04 & 41.80 & 42.44 \\
\midrule
Qwen3.5-4B & 42.68 & 73.72 & 30.00 & 67.20 & 38.68 & 38.64 & 35.60 & 37.48 & 30.72 & 32.80 & 35.68 & 38.08 & 45.28 & 29.40 & 34.16 & 39.00 & 37.20 & 41.28 & 40.42 \\
AfriqueQwen3.5-4B & 47.44 & 67.60 & 27.28 & 57.68 & 45.52 & 42.88 & 36.36 & 34.48 & 33.20 & 43.40 & 41.80 & 46.40 & 51.52 & 29.92 & 30.96 & 40.80 & 40.16 & 44.00 & 42.30 \\
AfriqueQwen3.5-4B-ExtendedCM & 48.64 & 69.08 & 31.96 & 59.76 & 44.96 & 43.32 & 39.92 & 36.32 & 34.08 & 45.92 & 43.08 & 47.32 & 52.20 & 30.00 & 31.24 & 45.04 & 42.32 & 45.12 & 43.90 \\
AfriqueQwen3.5-4B-50Langs & 51.36 & 68.76 & 32.96 & 59.68 & 46.16 & 45.20 & 39.44 & 41.72 & 40.08 & 43.60 & 43.24 & 46.32 & 50.76 & 38.12 & 35.36 & 45.20 & 41.24 & 45.04 & 45.24 \\
\midrule
Qwen3-8B & 40.96 & 77.80 & 33.56 & 69.68 & 34.48 & 35.68 & 32.08 & 41.16 & 31.40 & 37.64 & 36.28 & 35.60 & 42.00 & 33.28 & 33.60 & 34.00 & 36.80 & 36.60 & 40.14 \\
AfriqueQwen-8B & 56.32 & 78.12 & 31.00 & 67.80 & 48.20 & 45.76 & 40.92 & 39.44 & 33.32 & 46.44 & 45.76 & 47.24 & 52.00 & 31.24 & 31.04 & 43.88 & 43.52 & 46.00 & 46.00 \\
\midrule
Qwen3-14B & 45.36 & 82.40 & 34.64 & 73.00 & 37.00 & 35.56 & 37.36 & 41.64 & 34.48 & 39.76 & 36.48 & 38.92 & 47.04 & 32.96 & 33.12 & 37.96 & 42.28 & 38.56 & 42.70 \\
\rowcolor[HTML]{E8F5E9}
AfriqueQwen-14B & 59.44 & 80.68 & 32.96 & 72.92 & 52.80 & 49.96 & 46.12 & 39.84 & 36.12 & 52.68 & 48.68 & 54.08 & 61.56 & 31.68 & 31.20 & 51.36 & 47.28 & 50.44 & 49.99 \\

\bottomrule
\end{tabular}
\end{adjustbox}
\caption{5-shot performance on the AfriMMLU benchmark for massive multilingual language understanding. (MMLU)}
\label{tab:afrimmlu}
\end{table*}

\begin{table*}[h]
\centering

\begin{adjustbox}{width=\textwidth}
\begin{tabular}{lccccccccccccccccccc}
\toprule
model & amh & eng & ewe & fra & hau & ibo & kin & lin & lug & orm & sna & sot & swa & twi & wol & xho & yor & zul & avg \\
\midrule
Llama3.1-8B & 37.37 & 52.80 & 34.83 & 50.03 & 40.77 & 39.83 & 35.23 & 34.50 & 37.73 & 37.10 & 37.73 & 37.80 & 41.27 & 35.93 & 34.67 & 35.63 & 37.57 & 36.63 & 38.75 \\
Lugha-Llama-8B-wura & 38.17 & 50.23 & 35.07 & 48.37 & 40.20 & 41.17 & 35.83 & 34.33 & 36.90 & 39.57 & 38.33 & 40.10 & 41.20 & 35.23 & 34.47 & 39.73 & 39.03 & 38.27 & 39.23 \\
AfriqueLlama-8B & 37.17 & 43.80 & 32.70 & 42.10 & 37.13 & 38.13 & 35.93 & 32.57 & 34.97 & 37.73 & 36.87 & 37.83 & 37.80 & 33.07 & 31.87 & 39.10 & 36.93 & 36.63 & 36.80 \\
\midrule
Gemma3-4B & 38.83 & 47.10 & 34.67 & 44.07 & 39.43 & 38.40 & 36.77 & 34.17 & 34.67 & 35.03 & 37.67 & 37.13 & 40.13 & 32.93 & 33.40 & 38.57 & 36.67 & 36.73 & 37.58 \\
AfriqueGemma-4B & 39.90 & 44.97 & 34.07 & 44.07 & 40.90 & 40.23 & 37.97 & 33.43 & 36.27 & 37.00 & 40.43 & 42.17 & 40.87 & 33.77 & 33.37 & 39.43 & 38.97 & 37.93 & 38.65 \\
\midrule
Gemma3-12B & 43.23 & 58.07 & 34.30 & 55.23 & 47.70 & 44.87 & 39.70 & 32.43 & 42.63 & 41.97 & 45.80 & 44.80 & 48.73 & 36.27 & 33.33 & 44.90 & 41.90 & 40.50 & 43.13 \\
AfriqueGemma-12B & 43.47 & 54.20 & 34.10 & 50.67 & 47.27 & 45.57 & 38.47 & 32.80 & 41.43 & 45.93 & 46.00 & 46.07 & 46.60 & 35.43 & 33.83 & 45.30 & 45.03 & 40.93 & 42.95 \\
\midrule
Qwen3-4B & 38.90 & 62.73 & 35.50 & 59.77 & 35.77 & 36.80 & 34.47 & 33.73 & 36.53 & 39.30 & 37.00 & 35.83 & 42.57 & 35.27 & 33.07 & 36.67 & 36.30 & 34.73 & 39.16 \\
AfriqueQwen-4B & 44.33 & 56.43 & 33.00 & 52.00 & 46.83 & 46.57 & 37.60 & 32.17 & 37.80 & 46.00 & 46.57 & 47.80 & 46.43 & 33.90 & 32.37 & 45.50 & 44.63 & 41.43 & 42.85 \\
\midrule
Qwen3.5-4B & 42.17 & 55.13 & 33.80 & 53.60 & 40.40 & 43.80 & 37.40 & 32.03 & 35.90 & 38.00 & 39.27 & 42.10 & 42.00 & 34.53 & 35.23 & 40.27 & 40.33 & 38.27 & 40.24 \\
AfriqueQwen3.5-4B & 40.90 & 53.03 & 32.57 & 47.67 & 42.63 & 40.90 & 36.93 & 32.17 & 37.17 & 42.00 & 41.50 & 42.47 & 42.40 & 33.03 & 32.87 & 42.43 & 40.13 & 39.30 & 40.01 \\
AfriqueQwen3.5-4B-ExtendedCM & 41.87 & 54.50 & 32.80 & 50.47 & 43.73 & 43.10 & 35.27 & 31.33 & 38.13 & 42.33 & 42.67 & 44.83 & 43.53 & 34.53 & 31.90 & 42.33 & 41.43 & 40.20 & 40.83 \\
AfriqueQwen3.5-4B-50Langs & 42.67 & 54.37 & 35.73 & 50.90 & 44.37 & 43.37 & 34.57 & 33.17 & 41.47 & 41.67 & 41.13 & 44.60 & 44.87 & 35.90 & 35.27 & 41.90 & 41.13 & 39.40 & 41.47 \\
\midrule
Qwen3-8B & 40.83 & 62.77 & 34.03 & 61.93 & 35.83 & 38.87 & 34.90 & 32.50 & 35.63 & 38.80 & 37.73 & 37.27 & 45.97 & 34.63 & 33.20 & 37.50 & 38.17 & 34.77 & 39.74 \\
AfriqueQwen-8B & 44.63 & 60.67 & 32.43 & 58.20 & 48.03 & 45.13 & 39.57 & 32.47 & 38.83 & 50.07 & 47.47 & 48.50 & 50.27 & 33.37 & 31.90 & 46.33 & 43.93 & 41.93 & 44.10 \\
\midrule
Qwen3-14B & 43.70 & 66.10 & 35.93 & 64.10 & 43.03 & 43.33 & 37.63 & 32.80 & 38.40 & 45.07 & 43.37 & 42.87 & 50.30 & 35.87 & 33.60 & 41.13 & 45.43 & 39.60 & 43.46 \\
\rowcolor[HTML]{E8F5E9}
AfriqueQwen-14B & 48.77 & 60.60 & 34.87 & 58.80 & 49.90 & 49.87 & 41.40 & 34.03 & 42.90 & 51.07 & 51.70 & 49.20 & 52.57 & 34.53 & 32.10 & 48.57 & 49.73 & 46.33 & 46.50 \\

\bottomrule
\end{tabular}
\end{adjustbox}
\caption{5-shot performance on the AfriXNLI benchmark for cross-lingual natural language inference. (NLI)}
\label{tab:afrixnli}
\end{table*}

\begin{table*}[h]
\centering

\begin{adjustbox}{width=\textwidth}
\begin{tabular}{lcccccccccccccccccccccccccc}
\toprule
model & afr & amh & ary & arz & eng & fra & hau & ibo & kin & lin & lug & nya & orm & plt & por & sna & som & sot & swa & tir & tsn & wol & xho & yor & zul & avg \\
\midrule
Llama3.1-8B & 77.96 & 35.62 & 54.00 & 63.53 & 87.64 & 82.04 & 44.22 & 38.82 & 37.76 & 33.80 & 33.71 & 31.60 & 31.87 & 43.98 & 82.07 & 36.47 & 32.71 & 31.36 & 53.22 & 31.36 & 33.60 & 29.60 & 34.42 & 30.69 & 34.87 & 45.08 \\
Lugha-Llama-8B-wura & 79.16 & 47.40 & 51.11 & 62.47 & 84.51 & 79.87 & 53.98 & 42.13 & 44.60 & 34.58 & 34.18 & 41.18 & 37.04 & 57.98 & 79.60 & 47.04 & 45.89 & 40.44 & 61.96 & 35.80 & 39.16 & 27.89 & 42.82 & 35.42 & 43.82 & 49.75 \\
AfriqueLlama-8B & 71.31 & 54.62 & 54.49 & 59.69 & 79.31 & 73.78 & 49.16 & 41.49 & 49.51 & 31.87 & 33.80 & 43.69 & 42.00 & 58.09 & 73.47 & 50.31 & 46.69 & 47.40 & 61.53 & 43.56 & 47.96 & 27.13 & 48.60 & 41.40 & 48.24 & 51.16 \\
\midrule
Gemma3-4B & 73.64 & 52.73 & 51.51 & 61.71 & 78.58 & 76.00 & 47.24 & 36.16 & 43.07 & 30.91 & 33.53 & 41.44 & 31.16 & 54.62 & 73.24 & 44.33 & 41.93 & 37.78 & 63.80 & 35.31 & 35.78 & 28.09 & 40.44 & 32.71 & 44.58 & 47.61 \\
AfriqueGemma-4B & 67.93 & 55.93 & 51.56 & 56.87 & 74.96 & 68.89 & 51.24 & 41.00 & 50.78 & 30.29 & 33.40 & 46.11 & 40.64 & 58.44 & 68.02 & 52.40 & 48.11 & 49.02 & 60.67 & 45.29 & 48.18 & 26.04 & 48.18 & 39.69 & 47.89 & 50.46 \\
\midrule
Gemma3-12B & 90.71 & 76.98 & 77.98 & 82.53 & 92.53 & 90.33 & 73.36 & 55.80 & 70.91 & 43.02 & 49.07 & 63.07 & 52.33 & 79.18 & 89.69 & 71.00 & 69.33 & 64.93 & 86.24 & 51.27 & 55.73 & 32.38 & 67.82 & 50.36 & 68.49 & 68.20 \\
AfriqueGemma-12B & 86.29 & 74.71 & 70.87 & 76.11 & 89.18 & 85.96 & 68.58 & 55.87 & 70.89 & 35.36 & 44.93 & 62.76 & 60.29 & 79.69 & 84.80 & 69.53 & 66.16 & 67.76 & 79.11 & 62.42 & 65.11 & 28.89 & 66.09 & 55.27 & 66.87 & 66.94 \\
\midrule
Qwen3-4B & 83.60 & 44.82 & 56.80 & 74.36 & 90.33 & 88.78 & 32.44 & 31.36 & 36.71 & 34.11 & 31.22 & 32.47 & 37.29 & 39.87 & 85.33 & 37.98 & 33.47 & 31.78 & 50.42 & 32.67 & 33.24 & 32.98 & 32.96 & 31.78 & 34.51 & 46.05 \\
AfriqueQwen-4B & 84.58 & 71.56 & 68.13 & 76.00 & 89.67 & 87.49 & 62.76 & 51.67 & 61.04 & 32.98 & 35.93 & 56.09 & 54.38 & 72.58 & 85.91 & 65.27 & 56.78 & 59.16 & 78.20 & 61.04 & 57.49 & 30.13 & 61.80 & 49.13 & 61.11 & 62.84 \\
\midrule
Qwen3.5-4B & 87.29 & 60.73 & 66.96 & 76.76 & 92.78 & 90.64 & 52.11 & 48.53 & 58.31 & 38.18 & 36.29 & 44.80 & 40.38 & 53.58 & 90.04 & 50.44 & 53.82 & 47.27 & 73.18 & 49.27 & 49.04 & 36.22 & 52.64 & 43.38 & 52.18 & 57.79 \\
AfriqueQwen3.5-4B & 86.87 & 73.80 & 67.29 & 77.58 & 89.91 & 86.29 & 66.60 & 52.71 & 63.87 & 34.07 & 41.80 & 59.16 & 55.76 & 73.60 & 85.40 & 65.51 & 63.69 & 62.49 & 79.13 & 63.80 & 61.16 & 31.60 & 64.64 & 54.71 & 61.87 & 64.93 \\
AfriqueQwen3.5-4B-ExtendedCM & 88.02 & 74.09 & 67.04 & 77.13 & 90.40 & 86.16 & 66.20 & 56.24 & 66.18 & 35.18 & 44.00 & 59.44 & 56.49 & 74.67 & 86.67 & 65.11 & 62.27 & 64.42 & 77.53 & 63.69 & 62.27 & 31.11 & 65.04 & 54.13 & 62.58 & 65.44 \\
AfriqueQwen3.5-4B-50Langs & 87.53 & 74.38 & 67.27 & 75.98 & 89.84 & 86.80 & 68.02 & 55.38 & 64.58 & 55.71 & 54.58 & 60.98 & 56.31 & 73.82 & 86.49 & 66.80 & 63.96 & 64.93 & 78.16 & 63.22 & 63.04 & 44.40 & 65.82 & 54.98 & 64.58 & 67.50 \\
\midrule
Qwen3-8B & 87.96 & 50.24 & 64.04 & 79.69 & 91.67 & 90.04 & 32.69 & 34.78 & 37.67 & 35.53 & 31.44 & 32.73 & 38.13 & 44.20 & 88.13 & 40.20 & 33.78 & 35.82 & 62.67 & 35.04 & 35.02 & 33.07 & 35.04 & 31.71 & 36.53 & 48.71 \\
AfriqueQwen-8B & 88.60 & 75.18 & 73.29 & 79.73 & 91.76 & 90.11 & 67.20 & 54.11 & 67.76 & 35.76 & 39.49 & 61.64 & 60.64 & 76.96 & 88.56 & 67.40 & 63.51 & 65.49 & 81.31 & 67.18 & 63.29 & 32.44 & 66.67 & 55.89 & 64.98 & 67.16 \\
\midrule
Qwen3-14B & 90.78 & 55.42 & 72.27 & 84.78 & 94.56 & 93.16 & 39.87 & 39.29 & 42.31 & 38.42 & 36.58 & 38.44 & 41.69 & 55.31 & 91.53 & 43.93 & 35.13 & 43.69 & 73.33 & 39.53 & 41.27 & 33.82 & 44.18 & 37.44 & 45.40 & 54.09 \\
\rowcolor[HTML]{E8F5E9}
AfriqueQwen-14B & 91.04 & 82.33 & 78.47 & 84.69 & 93.40 & 92.22 & 75.49 & 62.71 & 72.67 & 38.02 & 45.38 & 68.24 & 66.24 & 82.62 & 90.93 & 73.56 & 69.49 & 73.82 & 85.58 & 74.09 & 69.11 & 32.87 & 72.51 & 62.80 & 72.58 & 72.43 \\
\bottomrule
\end{tabular}
\end{adjustbox}
\caption{5-shot performance on the Belebele reading comprehension benchmark. (RC)}
\label{tab:belebele}
\end{table*}

\begin{table*}[h]
\centering

\begin{adjustbox}{width=\textwidth}
\begin{tabular}{lccccccccccccccccccccccccc}
\toprule
model & aeb & afr & amh & ary & arz & ewe & hau & ibo & kin & lin & lug & nya & orm & sna & som & sot & swa & tir & tsn & twi & wol & xho & yor & zul & avg \\
\midrule
Llama3.1-8B & 55.19 & 69.24 & 4.81 & 49.39 & 59.42 & 25.57 & 31.31 & 25.63 & 12.19 & 14.88 & 20.92 & 21.63 & 19.83 & 17.93 & 15.55 & 16.34 & 46.08 & 7.87 & 15.79 & 16.16 & 32.89 & 12.46 & 16.45 & 9.91 & 25.73 \\
Lugha-Llama-8B-wura & 52.40 & 72.45 & 31.70 & 46.90 & 56.35 & 26.67 & 57.88 & 49.64 & 51.92 & 15.13 & 30.44 & 59.67 & 35.46 & 52.49 & 52.79 & 51.99 & 64.48 & 24.50 & 45.87 & 14.12 & 34.89 & 46.32 & 46.23 & 48.99 & 44.55 \\
AfriqueLlama-8B & 57.71 & 73.92 & 55.05 & 55.46 & 66.12 & 26.02 & 63.70 & 60.93 & 67.13 & 17.57 & 41.73 & 72.06 & 62.95 & 67.66 & 61.55 & 68.79 & 71.43 & 49.05 & 63.82 & 14.22 & 34.59 & 63.71 & 62.68 & 64.71 & 55.94 \\
\midrule
Gemma3-4B & 44.54 & 58.65 & 34.70 & 38.49 & 47.01 & 29.85 & 38.33 & 29.41 & 31.35 & 23.55 & 36.95 & 44.25 & 30.24 & 38.39 & 35.57 & 32.20 & 51.96 & 10.89 & 29.57 & 17.77 & 36.19 & 33.65 & 9.82 & 32.86 & 34.01 \\
AfriqueGemma-4B & 27.56 & 73.01 & 48.56 & 27.61 & 42.12 & 25.30 & 59.32 & 58.60 & 65.54 & 20.10 & 37.04 & 69.45 & 56.20 & 66.15 & 58.83 & 38.80 & 65.73 & 47.25 & 56.23 & 15.42 & 31.98 & 62.59 & 57.19 & 63.24 & 48.91 \\
\midrule
Gemma3-12B & 48.88 & 59.16 & 42.32 & 45.82 & 50.47 & 28.46 & 43.11 & 47.15 & 46.71 & 27.34 & 36.47 & 54.16 & 32.21 & 51.43 & 45.38 & 51.29 & 51.52 & 24.09 & 38.24 & 23.75 & 40.90 & 37.09 & 26.11 & 42.58 & 41.44 \\
AfriqueGemma-12B & 58.97 & 74.69 & 60.20 & 58.28 & 66.83 & 31.23 & 63.74 & 61.79 & 68.40 & 26.77 & 50.68 & 73.03 & 64.98 & 68.16 & 61.74 & 69.44 & 72.41 & 53.41 & 64.85 & 25.69 & 44.37 & 64.73 & 64.21 & 65.80 & 58.93 \\
\midrule
Qwen3-4B & 54.18 & 63.10 & -1.13 & 46.90 & 57.49 & 28.03 & 9.25 & 8.55 & 9.50 & 15.04 & 17.87 & 18.23 & 18.78 & 15.33 & 8.44 & 14.13 & 12.03 & 9.11 & 15.14 & 14.23 & 33.08 & 8.96 & 7.82 & 7.30 & 20.47 \\
AfriqueQwen-4B & 56.12 & 72.63 & 51.97 & 51.19 & 62.86 & 27.79 & 61.14 & 57.54 & 62.84 & 15.33 & 29.71 & 68.93 & 58.50 & 64.10 & 58.64 & 65.55 & 67.52 & 36.78 & 61.12 & 13.05 & 34.69 & 59.44 & 59.13 & 60.61 & 52.38 \\
\midrule
Qwen3.5-4B & 58.32 & 69.06 & 14.64 & 48.94 & 62.48 & 26.28 & 24.97 & 26.13 & 24.79 & 16.55 & 20.24 & 25.59 & 20.56 & 23.76 & 26.58 & 29.55 & 40.08 & 9.45 & 24.35 & 11.84 & 36.08 & 29.97 & 21.02 & 28.92 & 30.01 \\
AfriqueQwen3.5-4B & 58.23 & 73.76 & 58.59 & 55.82 & 66.14 & 25.86 & 62.57 & 60.16 & 66.63 & 18.96 & 40.27 & 71.44 & 63.17 & 67.54 & 61.27 & 68.63 & 70.57 & 48.83 & 63.84 & 13.15 & 33.15 & 63.41 & 62.51 & 64.33 & 55.78 \\
AfriqueQwen3.5-4B-ExtendedCM & 57.75 & 73.56 & 58.12 & 55.35 & 65.98 & 26.86 & 63.04 & 60.22 & 66.78 & 18.17 & 41.28 & 71.79 & 63.64 & 67.48 & 61.39 & 68.49 & 70.81 & 47.46 & 63.98 & 12.40 & 32.79 & 63.68 & 62.95 & 64.17 & 55.76 \\
\rowcolor[HTML]{E8F5E9}
AfriqueQwen3.5-4B-50Langs & 61.47 & 75.53 & 60.31 & 58.89 & 67.97 & 52.95 & 62.80 & 60.77 & 67.88 & 60.17 & 61.35 & 71.74 & 62.53 & 67.80 & 62.80 & 67.80 & 70.89 & 50.56 & 64.43 & 52.18 & 55.02 & 64.38 & 63.22 & 64.92 & 62.85 \\
\midrule
Qwen3-8B & 57.60 & 67.60 & 1.26 & 51.22 & 62.06 & 28.09 & 7.22 & 9.75 & 7.20 & 14.07 & 18.52 & 18.01 & 17.81 & 15.30 & 6.51 & 14.66 & 19.76 & 7.83 & 14.94 & 12.87 & 31.92 & 9.15 & 6.57 & 6.99 & 21.12 \\
AfriqueQwen-8B & 57.57 & 73.24 & 55.75 & 54.35 & 65.07 & 28.20 & 62.19 & 59.42 & 64.94 & 15.85 & 32.91 & 70.86 & 61.68 & 66.34 & 60.29 & 67.47 & 70.04 & 44.03 & 62.87 & 15.07 & 35.06 & 61.97 & 60.70 & 62.70 & 54.52 \\
\midrule
Qwen3-14B & 58.12 & 69.95 & 6.56 & 51.93 & 62.55 & 27.69 & 9.94 & 12.19 & 8.30 & 18.63 & 20.04 & 19.42 & 20.14 & 15.71 & 9.97 & 16.62 & 32.46 & 6.42 & 15.63 & 16.13 & 34.88 & 13.50 & 10.09 & 11.77 & 23.69 \\
AfriqueQwen-14B & 57.58 & 73.92 & 58.44 & 55.74 & 66.41 & 26.59 & 63.62 & 61.06 & 66.98 & 19.06 & 41.12 & 72.04 & 63.97 & 67.61 & 61.92 & 68.85 & 71.19 & 46.41 & 64.66 & 15.25 & 34.77 & 63.65 & 62.69 & 64.91 & 56.19 \\
% \midrule
% Llama3.1-8B & 55.19 & 69.24 & 4.81 & 49.39 & 59.42 & 25.57 & 31.31 & 25.63 & 12.19 & 14.88 & 20.92 & 21.63 & 19.83 & 17.93 & 15.55 & 16.34 & 46.08 & 7.87 & 15.79 & 16.16 & 32.89 & 12.46 & 16.45 & 9.91 & 25.73 \\
% AfriqueLlama-8B & 57.71 & 73.92 & 55.05 & 55.46 & 66.12 & 26.02 & 63.70 & 60.93 & 67.13 & 17.57 & 41.73 & 72.06 & 62.95 & 67.66 & 61.55 & 68.79 & 71.43 & 49.05 & 63.82 & 14.22 & 34.59 & 63.71 & 62.68 & 64.71 & 55.94 \\
\bottomrule
\end{tabular}
\end{adjustbox}
\caption{Translation performance (SSA COMET score) from English to African languages on the FLORES-200 benchmark. (MT eng2xx)}
\label{tab:flores_eng2xx}
\end{table*}

\begin{table*}[h]
\centering

\begin{adjustbox}{width=\textwidth}
\begin{tabular}{lccccccccccccccccccccccccc}
\toprule
model & aeb & afr & amh & ary & arz & ewe & hau & ibo & kin & lin & lug & nya & orm & sna & som & sot & swa & tir & tsn & twi & wol & xho & yor & zul & avg \\
\midrule
Llama3.1-8B & 65.61 & 72.84 & 49.75 & 64.89 & 68.10 & 40.71 & 60.61 & 56.84 & 52.16 & 46.37 & 46.16 & 48.83 & 40.61 & 48.94 & 49.60 & 48.12 & 67.01 & 38.79 & 47.44 & 49.50 & 43.63 & 50.21 & 47.97 & 50.39 & 52.30 \\
Lugha-Llama-8B-wura & 64.93 & 73.37 & 63.17 & 63.72 & 67.45 & 39.84 & 66.26 & 61.80 & 64.36 & 45.50 & 50.36 & 63.50 & 52.76 & 63.31 & 63.11 & 64.84 & 69.77 & 55.45 & 61.13 & 45.28 & 42.53 & 64.74 & 56.75 & 66.25 & 59.59 \\
AfriqueLlama-8B & 68.27 & 73.47 & 67.01 & 67.07 & 69.17 & 39.23 & 67.19 & 64.21 & 67.19 & 45.65 & 52.90 & 65.54 & 61.00 & 66.00 & 64.93 & 68.51 & 70.34 & 61.86 & 65.76 & 45.95 & 40.64 & 67.60 & 60.47 & 68.38 & 62.01 \\
\midrule
Gemma3-4B & 48.03 & 46.54 & 41.61 & 48.11 & 49.15 & 36.59 & 46.73 & 40.21 & 44.15 & 36.52 & 35.47 & 42.49 & 37.74 & 44.33 & 44.42 & 41.52 & 50.97 & 35.54 & 36.09 & 36.48 & 40.65 & 44.31 & 35.53 & 45.34 & 42.02 \\
AfriqueGemma-4B & 67.69 & 73.12 & 66.98 & 66.13 & 68.74 & 38.12 & 66.36 & 63.46 & 66.38 & 44.35 & 50.05 & 65.09 & 59.47 & 65.63 & 64.13 & 67.72 & 69.96 & 62.01 & 64.85 & 44.52 & 40.46 & 67.47 & 59.13 & 67.93 & 61.24 \\
\midrule
Gemma3-12B & 55.47 & 44.94 & 49.64 & 50.31 & 57.79 & 40.38 & 46.70 & 45.76 & 45.20 & 42.97 & 45.39 & 43.99 & 46.29 & 43.32 & 53.00 & 44.91 & 46.73 & 46.87 & 44.67 & 43.74 & 41.57 & 46.85 & 45.45 & 48.39 & 46.68 \\
AfriqueGemma-12B & 68.86 & 73.68 & 69.53 & 68.09 & 69.88 & 40.55 & 68.10 & 65.73 & 68.31 & 48.93 & 55.66 & 66.53 & 63.29 & 66.93 & 65.98 & 69.40 & 71.11 & 64.88 & 66.38 & 49.50 & 43.14 & 68.92 & 61.91 & 68.95 & 63.51 \\
\midrule
Qwen3-4B & 64.11 & 72.02 & 43.02 & 61.69 & 66.93 & 39.16 & 40.69 & 40.72 & 40.45 & 42.55 & 40.29 & 42.49 & 37.90 & 41.94 & 40.19 & 41.96 & 52.04 & 36.48 & 41.21 & 43.14 & 41.90 & 43.03 & 39.36 & 41.27 & 45.61 \\
AfriqueQwen-4B & 67.11 & 72.92 & 66.78 & 65.60 & 68.36 & 38.69 & 65.59 & 62.59 & 65.71 & 43.25 & 46.97 & 64.81 & 58.78 & 65.23 & 62.98 & 67.41 & 69.23 & 61.90 & 64.39 & 42.44 & 40.56 & 66.92 & 58.99 & 67.45 & 60.61 \\
\midrule
Qwen3.5-4B & 65.77 & 72.71 & 58.97 & 64.68 & 68.09 & 40.92 & 57.41 & 57.39 & 59.91 & 46.20 & 44.81 & 55.23 & 40.98 & 54.42 & 58.28 & 57.62 & 66.13 & 51.15 & 53.88 & 45.58 & 46.69 & 59.37 & 50.37 & 60.41 & 55.71 \\
AfriqueQwen3.5-4B & 68.40 & 73.37 & 68.39 & 67.27 & 69.04 & 39.23 & 67.30 & 64.77 & 67.29 & 46.82 & 54.05 & 65.85 & 61.60 & 66.10 & 64.92 & 68.98 & 70.52 & 64.11 & 66.14 & 44.74 & 40.89 & 68.07 & 61.16 & 68.64 & 62.40 \\
AfriqueQwen3.5-4B-ExtendedCM & 68.29 & 73.45 & 68.36 & 67.15 & 69.11 & 38.85 & 67.27 & 64.65 & 67.20 & 46.22 & 53.82 & 65.94 & 61.99 & 66.22 & 65.12 & 69.03 & 70.47 & 64.25 & 66.34 & 45.10 & 40.86 & 68.05 & 61.22 & 68.69 & 62.40 \\
\rowcolor[HTML]{E8F5E9} 
AfriqueQwen3.5-4B-50Langs & 66.64 & 75.07 & 67.05 & 65.54 & 69.07 & 56.50 & 66.31 & 61.51 & 66.20 & 63.27 & 61.42 & 66.86 & 60.97 & 66.41 & 64.96 & 68.72 & 71.06 & 62.94 & 66.13 & 58.06 & 58.14 & 67.20 & 61.58 & 67.61 & 64.97 \\
\midrule
Qwen3-8B & 65.66 & 72.86 & 50.20 & 64.00 & 68.46 & 39.18 & 41.41 & 43.09 & 42.15 & 43.99 & 41.27 & 43.86 & 39.81 & 43.64 & 41.58 & 44.33 & 59.72 & 40.44 & 42.79 & 43.78 & 42.24 & 46.31 & 41.67 & 44.34 & 47.78 \\
AfriqueQwen-8B & 68.52 & 73.47 & 68.31 & 67.21 & 69.45 & 39.24 & 67.06 & 64.07 & 66.92 & 44.64 & 49.61 & 65.39 & 61.25 & 66.08 & 64.67 & 68.44 & 70.13 & 63.94 & 65.75 & 44.10 & 41.44 & 68.07 & 60.68 & 68.13 & 61.94 \\
\midrule
Qwen3-14B & 67.15 & 73.34 & 52.57 & 65.77 & 69.56 & 40.68 & 44.45 & 46.68 & 44.80 & 45.68 & 42.96 & 46.61 & 43.22 & 45.47 & 43.46 & 48.09 & 63.94 & 43.02 & 45.18 & 45.69 & 43.72 & 50.32 & 45.04 & 48.89 & 50.26 \\
AfriqueQwen-14B & 69.25 & 73.78 & 69.57 & 68.11 & 70.13 & 40.49 & 67.99 & 65.48 & 67.75 & 46.87 & 53.12 & 66.47 & 63.12 & 66.98 & 65.59 & 69.23 & 70.93 & 65.04 & 66.53 & 45.78 & 42.58 & 68.85 & 62.37 & 69.14 & 63.13 \\
% \midrule
% Llama3.1-8B & 65.61 & 72.84 & 49.75 & 64.89 & 68.10 & 40.71 & 60.61 & 56.84 & 52.16 & 46.37 & 46.16 & 48.83 & 40.61 & 48.94 & 49.60 & 48.12 & 67.01 & 38.79 & 47.44 & 49.50 & 43.63 & 50.21 & 47.97 & 50.39 & 52.30 \\
% AfriqueLlama-8B & 68.27 & 73.47 & 67.01 & 67.07 & 69.17 & 39.23 & 67.19 & 64.21 & 67.19 & 45.65 & 52.90 & 65.54 & 61.00 & 66.00 & 64.93 & 68.51 & 70.34 & 61.86 & 65.76 & 45.95 & 40.64 & 67.60 & 60.47 & 68.38 & 62.01 \\
\bottomrule
\end{tabular}
\end{adjustbox}
\caption{Translation performance (SSA COMET score) from African languages to English on the FLORES-200 benchmark. (xx2eng)}
\label{tab:flores_xx2eng}
\end{table*}

\begin{table*}[h]
\centering
\begin{adjustbox}{width=\textwidth}
\begin{tabular}{lcccccccccccccccccccccccccc}
\toprule
\multicolumn{26}{c}{\textit{English $\rightarrow$ African Languages (eng2xx) -- chrF++}} \\
\midrule
model & aeb & afr & amh & ary & arz & ewe & hau & ibo & kin & lin & lug & nya & orm & sna & som & sot & swa & tir & tsn & twi & wol & xho & yor & zul & avg \\
\midrule
Llama3.1-8B & 33.87 & 62.46 & 7.43 & 28.37 & 34.66 & 6.64 & 28.59 & 19.42 & 12.22 & 13.12 & 10.98 & 12.51 & 10.79 & 11.40 & 17.38 & 13.25 & 37.69 & 4.12 & 14.83 & 15.92 & 9.14 & 12.36 & 12.72 & 12.41 & 18.43 \\
Lugha-Llama-8B-wura & 31.45 & 65.35 & 16.95 & 26.26 & 31.93 & 8.46 & 43.80 & 32.25 & 30.56 & 12.66 & 15.46 & 33.13 & 17.79 & 30.31 & 34.29 & 35.92 & 49.79 & 10.47 & 32.01 & 12.11 & 9.58 & 30.71 & 20.27 & 33.54 & 27.71 \\
AfriqueLlama-8B & 36.56 & 67.54 & 32.11 & 32.43 & 40.17 & 6.80 & 49.56 & 40.84 & 46.44 & 14.15 & 20.96 & 45.64 & 35.54 & 42.87 & 41.95 & 53.84 & 59.54 & 19.80 & 46.14 & 13.89 & 8.93 & 47.37 & 27.36 & 49.67 & 36.67 \\
\midrule
Gemma3-4B & 17.89 & 39.24 & 12.07 & 12.10 & 16.98 & 7.73 & 26.01 & 20.91 & 12.34 & 12.00 & 7.32 & 14.98 & 7.94 & 14.51 & 15.91 & 14.90 & 31.09 & 3.11 & 11.63 & 11.25 & 7.67 & 14.01 & 7.62 & 18.15 & 14.89 \\
AfriqueGemma-4B & 5.49 & 66.53 & 19.88 & 6.10 & 12.10 & 8.74 & 44.07 & 39.29 & 44.72 & 11.91 & 18.34 & 42.05 & 27.29 & 40.88 & 37.94 & 16.82 & 51.14 & 14.25 & 34.62 & 10.28 & 8.11 & 46.21 & 23.04 & 47.49 & 28.22 \\
\midrule
Gemma3-12B & 25.47 & 43.57 & 20.81 & 21.54 & 23.40 & 7.45 & 24.79 & 32.27 & 22.03 & 14.16 & 10.18 & 17.14 & 11.86 & 22.58 & 20.58 & 29.82 & 28.01 & 6.38 & 20.09 & 15.11 & 7.60 & 9.21 & 13.91 & 17.76 & 19.41 \\
AfriqueGemma-12B & 39.54 & 68.97 & 36.09 & 36.24 & 41.66 & 8.47 & 51.36 & 41.95 & 49.58 & 19.64 & 26.60 & 46.87 & 36.98 & 43.94 & 42.14 & 55.54 & 62.13 & 20.90 & 46.61 & 15.58 & 7.11 & 49.21 & 28.59 & 51.82 & 38.65 \\
\midrule
Qwen3-4B & 32.58 & 55.22 & 3.45 & 26.71 & 32.68 & 6.29 & 7.66 & 7.15 & 7.01 & 9.80 & 7.07 & 8.03 & 7.80 & 7.23 & 9.63 & 9.24 & 14.19 & 2.36 & 8.29 & 8.90 & 7.90 & 7.40 & 5.60 & 7.02 & 12.47 \\
AfriqueQwen-4B & 33.40 & 65.24 & 28.84 & 28.48 & 36.37 & 6.74 & 46.97 & 37.85 & 41.07 & 10.70 & 14.30 & 42.32 & 31.90 & 39.41 & 39.24 & 49.15 & 54.33 & 15.24 & 43.16 & 9.12 & 7.55 & 42.03 & 25.84 & 44.02 & 33.05 \\
\midrule
Qwen3.5-4B & 36.73 & 60.29 & 11.97 & 28.21 & 38.16 & 8.18 & 25.26 & 19.80 & 17.21 & 12.31 & 9.33 & 14.39 & 10.44 & 15.37 & 22.19 & 21.58 & 35.68 & 5.58 & 20.02 & 11.64 & 11.93 & 23.08 & 13.87 & 24.26 & 20.73 \\
AfriqueQwen3.5-4B & 37.68 & 67.30 & 33.15 & 33.30 & 41.38 & 6.90 & 50.08 & 40.94 & 46.43 & 13.73 & 20.28 & 45.23 & 35.41 & 42.78 & 42.15 & 53.69 & 59.49 & 19.15 & 45.80 & 13.11 & 7.12 & 47.21 & 28.23 & 50.21 & 36.70 \\
AfriqueQwen3.5-4B-ExtendedCM & 37.20 & 67.23 & 33.36 & 32.76 & 41.27 & 6.85 & 49.96 & 41.14 & 46.69 & 13.21 & 20.73 & 45.62 & 36.13 & 42.83 & 42.25 & 54.02 & 59.95 & 18.87 & 45.99 & 12.57 & 6.86 & 47.47 & 28.38 & 50.20 & 36.73 \\
\rowcolor[HTML]{E8F5E9}
AfriqueQwen3.5-4B-50Langs & 37.67 & 67.07 & 33.07 & 32.65 & 41.13 & 31.03 & 49.92 & 40.92 & 46.39 & 45.23 & 34.57 & 45.27 & 35.80 & 43.00 & 42.27 & 53.98 & 59.61 & 19.22 & 46.04 & 31.87 & 25.79 & 47.00 & 27.51 & 50.35 & 41.14 \\
\midrule
Qwen3-8B & 35.19 & 59.93 & 6.46 & 29.63 & 36.23 & 5.67 & 8.72 & 9.08 & 7.34 & 10.53 & 6.90 & 8.41 & 8.80 & 6.87 & 11.19 & 10.08 & 20.56 & 2.49 & 9.44 & 9.00 & 7.74 & 10.37 & 7.65 & 9.73 & 14.08 \\
AfriqueQwen-8B & 35.86 & 66.94 & 31.84 & 31.29 & 39.02 & 6.71 & 48.40 & 39.82 & 43.70 & 12.04 & 16.75 & 44.42 & 34.11 & 41.43 & 40.79 & 51.64 & 57.72 & 17.88 & 44.93 & 9.58 & 8.71 & 44.66 & 27.21 & 46.92 & 35.10 \\
\midrule
Qwen3-14B & 36.03 & 61.78 & 9.03 & 30.45 & 37.88 & 6.26 & 13.45 & 11.98 & 10.51 & 15.09 & 8.83 & 10.67 & 11.82 & 9.82 & 15.17 & 13.97 & 31.17 & 3.26 & 11.68 & 13.49 & 10.23 & 13.91 & 10.25 & 14.08 & 17.12 \\
AfriqueQwen-14B & 36.80 & 67.58 & 33.52 & 32.50 & 40.73 & 6.22 & 49.78 & 41.14 & 46.52 & 14.67 & 21.20 & 45.92 & 36.39 & 42.90 & 41.62 & 53.50 & 59.55 & 18.38 & 46.15 & 12.83 & 8.41 & 46.50 & 27.45 & 49.82 & 36.67 \\
% \midrule
% Gemma3-27B & 34.85 & 59.41 & 27.52 & 29.65 & 34.04 & 4.71 & 21.76 & 35.10 & 17.54 & 19.80 & 15.35 & 6.81 & 18.16 & 10.04 & 19.42 & 27.30 & 23.05 & 10.42 & 15.68 & 17.15 & 6.98 & 25.95 & 18.31 & 21.10 & 21.67 \\
\bottomrule
\end{tabular}
\end{adjustbox}
\caption{Translation performance (chrF++ score) from English to African languages on the FLORES-200 benchmark. (eng2xx)}
\label{tab:flores_chrf_eng2xx}
\end{table*}

\begin{table*}[h]
\centering
\begin{adjustbox}{width=\textwidth}
\begin{tabular}{lcccccccccccccccccccccccccc}
\toprule
\multicolumn{26}{c}{\textit{African Languages $\rightarrow$ English (xx2eng) -- chrF++}} \\
\midrule
model & aeb & afr & amh & ary & arz & ewe & hau & ibo & kin & lin & lug & nya & orm & sna & som & sot & swa & tir & tsn & twi & wol & xho & yor & zul & avg \\
\midrule
Llama3.1-8B & 52.59 & 74.19 & 31.70 & 50.96 & 55.37 & 21.93 & 45.81 & 40.79 & 34.24 & 27.55 & 28.07 & 30.36 & 22.14 & 30.05 & 31.31 & 30.80 & 56.85 & 20.68 & 29.32 & 30.82 & 25.20 & 32.93 & 29.78 & 33.25 & 36.11 \\
Lugha-Llama-8B-wura & 51.29 & 75.17 & 46.43 & 49.15 & 54.18 & 20.63 & 52.42 & 46.43 & 48.15 & 26.64 & 31.79 & 45.71 & 32.89 & 44.73 & 47.66 & 50.29 & 62.01 & 35.20 & 42.72 & 26.17 & 24.09 & 50.46 & 38.05 & 53.54 & 43.99 \\
AfriqueLlama-8B & 54.93 & 75.07 & 52.68 & 52.99 & 56.31 & 20.78 & 54.25 & 49.67 & 53.11 & 26.94 & 34.86 & 48.95 & 43.42 & 48.09 & 50.83 & 56.86 & 63.08 & 42.92 & 48.56 & 27.20 & 22.37 & 55.89 & 42.43 & 57.62 & 47.49 \\
\midrule
Gemma3-4B & 21.81 & 20.65 & 19.13 & 22.48 & 22.79 & 10.85 & 25.36 & 18.63 & 17.60 & 13.16 & 11.93 & 16.19 & 13.75 & 20.11 & 24.62 & 17.51 & 30.95 & 14.05 & 14.31 & 16.89 & 12.89 & 21.26 & 14.77 & 23.61 & 18.55 \\
AfriqueGemma-4B & 54.91 & 74.89 & 53.81 & 51.94 & 55.94 & 18.83 & 53.22 & 48.91 & 52.44 & 25.84 & 32.11 & 48.56 & 41.75 & 48.00 & 50.04 & 55.96 & 62.70 & 43.61 & 47.74 & 25.90 & 21.38 & 55.84 & 41.40 & 57.12 & 46.79 \\
\midrule
Gemma3-12B & 32.42 & 11.50 & 21.56 & 21.58 & 35.81 & 13.92 & 15.27 & 15.07 & 11.51 & 10.96 & 15.10 & 9.08 & 18.78 & 7.31 & 28.42 & 11.70 & 15.80 & 18.23 & 11.60 & 12.17 & 12.38 & 15.20 & 17.14 & 18.70 & 16.72 \\
AfriqueGemma-12B & 56.92 & 76.62 & 59.33 & 55.70 & 58.55 & 19.71 & 56.79 & 52.73 & 55.71 & 29.77 & 38.41 & 51.07 & 47.54 & 50.43 & 53.50 & 59.43 & 66.01 & 48.34 & 50.27 & 30.86 & 22.53 & 59.21 & 44.72 & 59.87 & 50.17 \\
\midrule
Qwen3-4B & 51.02 & 72.07 & 26.17 & 47.07 & 53.31 & 19.84 & 22.22 & 21.91 & 22.10 & 23.69 & 21.64 & 23.41 & 19.83 & 22.96 & 21.57 & 23.61 & 36.77 & 18.69 & 22.62 & 23.72 & 22.77 & 24.84 & 21.36 & 23.31 & 28.60 \\
AfriqueQwen-4B & 54.11 & 74.11 & 52.87 & 51.55 & 55.50 & 20.47 & 51.70 & 47.37 & 50.13 & 25.24 & 28.82 & 47.62 & 40.40 & 46.65 & 47.79 & 54.18 & 60.06 & 43.60 & 46.83 & 24.66 & 22.25 & 54.15 & 40.93 & 55.53 & 45.69 \\
\midrule
Qwen3.5-4B & 53.32 & 73.91 & 43.30 & 50.69 & 55.34 & 23.23 & 42.99 & 41.11 & 43.43 & 28.33 & 27.23 & 37.66 & 23.52 & 36.38 & 42.85 & 43.52 & 55.67 & 32.80 & 36.57 & 27.70 & 29.62 & 45.12 & 33.53 & 46.71 & 40.61 \\
AfriqueQwen3.5-4B & 56.04 & 75.16 & 55.84 & 53.73 & 56.61 & 20.74 & 54.51 & 50.85 & 53.34 & 28.33 & 36.52 & 49.27 & 44.77 & 48.77 & 51.42 & 57.91 & 63.40 & 46.79 & 49.88 & 26.86 & 22.64 & 56.77 & 43.96 & 58.44 & 48.44 \\
AfriqueQwen3.5-4B-ExtendedCM & 55.53 & 75.10 & 55.47 & 53.41 & 56.20 & 20.34 & 54.55 & 50.52 & 53.00 & 27.37 & 36.08 & 49.48 & 44.82 & 48.66 & 51.39 & 58.27 & 63.43 & 46.52 & 49.87 & 27.03 & 22.23 & 56.78 & 43.58 & 58.29 & 48.25 \\
\rowcolor[HTML]{E8F5E9}
AfriqueQwen3.5-4B-50Langs & 55.79 & 74.98 & 55.38 & 53.12 & 56.29 & 36.45 & 54.18 & 50.50 & 53.04 & 44.81 & 43.52 & 48.95 & 44.04 & 48.38 & 51.19 & 57.67 & 63.23 & 46.03 & 49.80 & 40.99 & 39.17 & 56.58 & 43.48 & 58.03 & 51.07 \\
\midrule
Qwen3-8B & 53.40 & 74.07 & 34.64 & 50.49 & 55.80 & 20.99 & 24.37 & 25.49 & 24.85 & 25.68 & 23.93 & 25.69 & 22.22 & 25.49 & 24.11 & 27.26 & 46.72 & 22.91 & 24.96 & 25.39 & 23.91 & 29.75 & 24.25 & 28.01 & 31.85 \\
AfriqueQwen-8B & 56.43 & 75.40 & 55.90 & 53.75 & 57.35 & 20.94 & 54.03 & 49.50 & 52.29 & 26.38 & 32.25 & 48.86 & 43.80 & 48.57 & 50.16 & 56.41 & 62.37 & 46.42 & 49.32 & 26.41 & 23.24 & 56.69 & 43.08 & 57.71 & 47.80 \\
\midrule
Qwen3-14B & 55.39 & 75.15 & 36.91 & 52.67 & 57.75 & 22.35 & 28.23 & 29.80 & 28.04 & 27.63 & 25.79 & 28.43 & 26.00 & 27.76 & 26.33 & 32.74 & 53.21 & 24.76 & 28.22 & 27.33 & 24.99 & 34.89 & 27.94 & 33.13 & 34.81 \\
AfriqueQwen-14B & 57.60 & 75.98 & 58.03 & 55.60 & 58.66 & 21.97 & 56.27 & 52.01 & 54.58 & 28.44 & 35.96 & 50.44 & 46.87 & 50.00 & 52.21 & 59.06 & 64.77 & 48.20 & 50.44 & 27.83 & 24.15 & 58.34 & 45.35 & 59.65 & 49.68 \\

% \midrule
% Gemma3-27B & 57.48 & 68.30 & 51.53 & 55.82 & 57.53 & 21.34 & 46.19 & 44.61 & 49.91 & 35.33 & 38.58 & 45.97 & 38.98 & 45.22 & 44.41 & 50.55 & 61.07 & 36.94 & 41.90 & 35.85 & 26.76 & 52.38 & 38.03 & 51.82 & 45.69 \\
\bottomrule
\end{tabular}
\end{adjustbox}
\caption{Translation performance (chrF++ score) from African languages to English on the FLORES-200 benchmark. (xx2eng)}
\label{tab:flores_chrf_xx2eng}
\end{table*}

\begin{table*}[h]
\centering

\begin{adjustbox}{width=\textwidth}
\begin{tabular}{lcccccccccccccccccc}
\toprule
model & amh & eng & ewe & hau & ibo & kin & lin & lug & orm & sna & sot & swa & twi & wol & xho & yor & zul & avg \\
\midrule
Llama3.1-8B & 42.72 & 83.18 & 12.47 & 60.94 & 53.63 & 31.78 & 40.53 & 32.56 & 18.12 & 28.97 & 28.41 & 75.00 & 32.31 & 27.27 & 38.81 & 42.44 & 34.22 & 40.20 \\
Lugha-Llama-8B-wura & 66.84 & 82.15 & 11.91 & 78.78 & 64.94 & 55.28 & 36.25 & 38.44 & 37.84 & 62.09 & 52.87 & 80.19 & 21.94 & 22.38 & 64.53 & 63.66 & 58.31 & 52.85 \\
AfriqueLlama-8B & 78.78 & 79.68 & 9.66 & 80.19 & 70.78 & 62.19 & 38.78 & 50.16 & 59.72 & 72.47 & 63.38 & 81.19 & 22.22 & 19.81 & 75.22 & 73.38 & 65.59 & 59.01 \\
\midrule
Gemma3-4B & 73.72 & 79.36 & 11.47 & 72.97 & 61.78 & 50.19 & 40.00 & 31.87 & 19.97 & 53.31 & 35.38 & 84.44 & 25.53 & 25.83 & 62.59 & 38.81 & 57.53 & 48.51 \\
AfriqueGemma-4B & 75.69 & 79.74 & 9.88 & 77.94 & 70.09 & 62.09 & 38.75 & 39.91 & 55.72 & 71.72 & 58.56 & 82.03 & 16.50 & 21.57 & 76.31 & 68.03 & 63.84 & 56.96 \\
\midrule
Gemma3-12B & 83.03 & 85.76 & 15.97 & 84.97 & 72.59 & 68.28 & 53.41 & 64.53 & 48.91 & 80.03 & 59.72 & 90.47 & 47.81 & 36.24 & 80.09 & 67.12 & 73.56 & 65.44 \\
AfriqueGemma-12B & 82.81 & 82.64 & 13.84 & 84.22 & 77.94 & 68.44 & 44.47 & 62.19 & 67.28 & 79.69 & 66.97 & 86.34 & 34.69 & 29.31 & 81.53 & 76.09 & 73.41 & 65.40 \\
\midrule
Qwen3-4B & 35.53 & 84.34 & 13.34 & 15.12 & 21.00 & 17.22 & 39.72 & 17.09 & 14.28 & 13.19 & 21.25 & 50.94 & 19.75 & 24.33 & 18.31 & 14.06 & 17.72 & 25.72 \\
AfriqueQwen-4B & 74.66 & 76.72 & 11.78 & 74.94 & 65.81 & 55.44 & 35.25 & 33.94 & 51.88 & 66.38 & 57.91 & 74.84 & 16.91 & 20.85 & 67.72 & 63.97 & 65.22 & 53.78 \\
\midrule
Qwen3.5-4B & 67.28 & 88.14 & 17.06 & 66.66 & 66.97 & 55.59 & 42.91 & 36.81 & 22.66 & 52.97 & 49.19 & 83.28 & 29.03 & 40.53 & 68.31 & 62.06 & 58.81 & 53.43 \\
AfriqueQwen3.5-4B & 81.25 & 81.67 & 13.19 & 86.84 & 77.69 & 64.59 & 42.50 & 58.31 & 64.69 & 75.97 & 65.75 & 85.28 & 25.16 & 26.33 & 78.94 & 78.91 & 70.19 & 63.37 \\
AfriqueQwen3.5-4B-ExtendedCM & 81.50 & 82.22 & 12.75 & 85.22 & 76.72 & 67.50 & 41.00 & 58.59 & 66.41 & 77.09 & 66.62 & 85.41 & 24.41 & 25.30 & 78.97 & 79.50 & 70.75 & 63.53 \\
\rowcolor[HTML]{E8F5E9}
AfriqueQwen3.5-4B-50Langs & 80.88 & 83.67 & 47.62 & 86.81 & 76.72 & 65.81 & 70.47 & 70.03 & 65.19 & 76.00 & 66.75 & 86.28 & 60.31 & 58.40 & 80.22 & 78.31 & 71.09 & 72.03 \\
\midrule
Qwen3-8B & 50.37 & 85.27 & 12.91 & 18.69 & 25.84 & 21.12 & 42.34 & 19.78 & 18.47 & 18.94 & 26.28 & 70.12 & 21.56 & 26.43 & 27.25 & 23.88 & 23.25 & 31.32 \\
AfriqueQwen-8B & 80.47 & 83.67 & 9.78 & 82.88 & 75.59 & 62.34 & 38.31 & 46.97 & 60.53 & 74.72 & 63.09 & 86.84 & 19.47 & 22.98 & 78.19 & 73.19 & 69.12 & 60.48 \\
\midrule
Qwen3-14B & 59.34 & 88.52 & 15.38 & 29.41 & 39.66 & 27.59 & 46.44 & 31.53 & 30.28 & 31.84 & 33.38 & 81.66 & 30.84 & 32.82 & 47.06 & 38.28 & 41.28 & 41.49 \\

AfriqueQwen-14B & 83.22 & 86.30 & 14.88 & 85.88 & 78.94 & 67.81 & 41.38 & 58.38 & 69.62 & 81.25 & 64.75 & 86.72 & 25.69 & 31.10 & 82.53 & 81.88 & 73.25 & 65.50 \\
\bottomrule
\end{tabular}
\end{adjustbox}
\caption{5-shot performance on the Injongo Intent Classification benchmark. (Intent)}
\label{tab:injongointent}
\end{table*}

\begin{table*}[h]
\centering

\begin{adjustbox}{width=\textwidth}
\begin{tabular}{lccccccccccccccccccccccccccc}
\toprule
model & aeb & afr & amh & ary & arz & eng & ewe & hau & ibo & kin & lin & lug & nya & orm & plt & por & sna & som & sot & swa & tir & twi & wol & xho & yor & zul & avg \\
\midrule
Llama3.1-8B & 76.19 & 81.24 & 49.91 & 80.06 & 80.90 & 80.78 & 44.72 & 67.67 & 67.39 & 54.60 & 47.94 & 45.20 & 62.61 & 42.00 & 60.97 & 82.43 & 47.57 & 58.33 & 50.27 & 66.50 & 41.63 & 62.87 & 46.26 & 52.36 & 47.44 & 52.28 & 59.62 \\
Lugha-Llama-8B-wura & 75.39 & 80.41 & 74.83 & 79.10 & 78.73 & 80.45 & 43.06 & 80.21 & 76.48 & 78.99 & 59.51 & 63.74 & 76.90 & 67.40 & 77.95 & 81.63 & 74.98 & 76.04 & 72.64 & 79.55 & 64.44 & 58.12 & 55.42 & 76.78 & 73.10 & 76.55 & 72.40 \\
AfriqueLlama-8B & 76.27 & 78.90 & 62.86 & 79.06 & 77.80 & 66.32 & 38.32 & 64.32 & 65.83 & 64.31 & 41.85 & 51.13 & 75.70 & 61.90 & 75.80 & 78.97 & 66.21 & 75.76 & 61.98 & 65.36 & 73.21 & 53.28 & 38.75 & 64.55 & 61.91 & 61.87 & 64.70 \\
\midrule
Gemma3-4B & 76.81 & 79.84 & 63.38 & 78.55 & 81.36 & 69.21 & 44.00 & 59.88 & 56.37 & 60.65 & 46.20 & 38.38 & 74.48 & 38.98 & 71.44 & 80.36 & 56.59 & 74.16 & 51.54 & 66.70 & 63.75 & 57.82 & 45.28 & 55.46 & 38.68 & 59.53 & 61.13 \\
AfriqueGemma-4B & 77.69 & 79.44 & 61.15 & 79.10 & 79.84 & 70.76 & 42.81 & 65.28 & 63.27 & 62.74 & 40.04 & 42.59 & 78.33 & 64.57 & 77.17 & 79.15 & 63.88 & 77.15 & 63.00 & 65.80 & 77.26 & 55.27 & 41.41 & 62.31 & 55.22 & 61.84 & 64.89 \\
\midrule
Gemma3-12B & 81.49 & 83.65 & 81.56 & 82.89 & 82.26 & 83.21 & 47.47 & 80.64 & 78.58 & 81.36 & 64.38 & 68.54 & 82.31 & 67.65 & 81.13 & 84.28 & 78.95 & 80.73 & 76.04 & 85.33 & 73.51 & 67.01 & 58.80 & 79.74 & 66.67 & 79.70 & 76.07 \\
AfriqueGemma-12B & 78.85 & 82.43 & 74.02 & 80.16 & 79.65 & 73.64 & 43.23 & 74.27 & 69.80 & 73.11 & 51.73 & 57.45 & 79.56 & 66.92 & 79.60 & 82.23 & 70.86 & 80.12 & 71.26 & 76.32 & 79.67 & 60.74 & 48.49 & 70.31 & 66.51 & 73.12 & 70.93 \\
\midrule
Qwen3-4B & 78.70 & 81.42 & 53.62 & 82.09 & 82.31 & 82.22 & 46.00 & 47.94 & 47.51 & 50.44 & 59.30 & 50.67 & 55.32 & 50.01 & 55.10 & 80.87 & 51.88 & 50.42 & 55.71 & 69.23 & 40.74 & 56.90 & 57.54 & 52.89 & 46.29 & 48.13 & 58.97 \\
AfriqueQwen-4B & 76.04 & 75.94 & 73.25 & 75.70 & 77.02 & 78.00 & 37.10 & 77.69 & 75.25 & 76.11 & 54.43 & 53.63 & 76.87 & 70.78 & 73.85 & 76.18 & 74.93 & 73.71 & 73.95 & 76.69 & 69.73 & 49.76 & 49.42 & 75.68 & 71.60 & 75.91 & 69.97 \\
\midrule
Qwen3.5-4B & 82.13 & 83.54 & 75.18 & 82.82 & 84.04 & 85.65 & 48.97 & 72.73 & 77.20 & 76.13 & 63.71 & 59.43 & 73.02 & 53.19 & 70.41 & 85.11 & 69.42 & 77.71 & 72.14 & 83.53 & 71.20 & 58.80 & 63.91 & 74.78 & 66.12 & 78.98 & 72.69 \\
AfriqueQwen3.5-4B & 79.22 & 81.86 & 79.99 & 80.72 & 80.40 & 82.48 & 42.49 & 81.45 & 76.33 & 78.15 & 63.23 & 65.86 & 80.45 & 78.82 & 79.42 & 82.03 & 79.43 & 79.97 & 79.89 & 82.59 & 78.33 & 56.09 & 55.64 & 80.04 & 74.98 & 81.41 & 75.43 \\
AfriqueQwen3.5-4B-ExtendedCM & 78.70 & 82.52 & 79.76 & 80.87 & 80.64 & 83.01 & 43.23 & 82.89 & 78.27 & 80.82 & 60.53 & 67.67 & 81.69 & 78.79 & 80.39 & 81.37 & 79.06 & 80.84 & 80.15 & 82.67 & 81.69 & 53.90 & 54.68 & 79.12 & 78.81 & 82.18 & 75.93 \\
\rowcolor[HTML]{E8F5E9}
AfriqueQwen3.5-4B-50Langs & 79.20 & 81.72 & 79.24 & 80.33 & 81.03 & 80.05 & 74.94 & 82.48 & 79.50 & 81 & 78.79 & 75.07 & 79.62 & 76.86 & 80.57 & 81.11 & 78.27 & 80.24 & 79.35 & 81.58 & 79.78 & 76.96 & 72.99 & 79.42 & 76.49 & 80.17 & 79.11 \\
\midrule
Qwen3-8B & 83.11 & 83.89 & 49.98 & 83.15 & 86.49 & 72.29 & 43.43 & 37.49 & 36.48 & 37.86 & 43.81 & 35.60 & 60.53 & 39.48 & 57.25 & 83.32 & 37.99 & 50.54 & 41.42 & 55.91 & 49.97 & 55.08 & 47.76 & 42.00 & 36.67 & 37.92 & 53.44 \\
AfriqueQwen-8B & 80.79 & 83.84 & 72.40 & 81.04 & 83.39 & 77.58 & 42.41 & 74.45 & 70.43 & 78.06 & 53.30 & 58.53 & 81.38 & 69.51 & 81.54 & 82.92 & 74.41 & 81.65 & 71.87 & 78.45 & 81.42 & 51.36 & 52.12 & 71.78 & 72.42 & 74.23 & 72.36 \\
\midrule
Qwen3-14B & 82.30 & 85.42 & 74.12 & 85.60 & 84.43 & 85.71 & 47.07 & 55.71 & 60.47 & 54.38 & 63.74 & 55.05 & 57.58 & 56.11 & 65.90 & 84.98 & 55.42 & 56.12 & 63.37 & 80.75 & 61.41 & 56.68 & 61.66 & 63.64 & 54.86 & 61.82 & 65.93 \\
AfriqueQwen-14B & 81.81 & 83.60 & 83.97 & 85.12 & 83.04 & 82.62 & 46.52 & 82.66 & 81.39 & 83.19 & 61.41 & 65.83 & 82.54 & 84.15 & 82.77 & 82.85 & 80.73 & 82.39 & 80.62 & 85.43 & 81.93 & 58.06 & 58.13 & 80.42 & 81.14 & 82.99 & 77.90 \\
\bottomrule
\end{tabular}
\end{adjustbox}
\caption{5-shot performance on the SIB-200 topic classification benchmark. (Topic)}
\label{tab:sib}
\end{table*}

\end{document}